\documentclass[journal]{IEEEtran}

\usepackage{subfigure} 
\usepackage{float} 
\usepackage{multirow} 
\usepackage{mathtools} 
\usepackage{amsmath} 
\usepackage{amssymb} 
\usepackage{verbatim} 
\usepackage[table]{xcolor} 
\usepackage{makecell} 
\usepackage[linesnumbered,ruled,vlined]{algorithm2e} 
\usepackage[referable]{threeparttablex} 
\usepackage{dblfloatfix} 

\ifCLASSINFOpdf
\else
\fi

\SetAlFnt{\footnotesize}
\SetAlCapFnt{\footnotesize}
\SetAlCapNameFnt{\footnotesize}
\usepackage{algorithmic}

\SetCommentSty{mycommfont}
\SetKwInput{KwInput}{Input} 
\SetKwInput{KwParams}{Parameters} 
\SetKwInput{KwOutput}{Output} 

\makeatletter
\newcommand{\removelatexerror}{\let\@latex@error\@gobble}
\makeatother

\usepackage{array}
\newcommand{\PreserveBackslash}[1]{\let\temp=\\#1\let\\=\temp}
\newcolumntype{C}[1]{>{\PreserveBackslash\centering}p{#1}}

\DeclareMathOperator{\atantwo}{atan2}
\newcommand{\pluseq}{\mathrel{+}=}
\newcommand{\minuseq}{\mathrel{-}=}
\newcommand{\argmin}[1]{\underset{#1}{\operatorname{arg}\,\operatorname{min}}\;}

\usepackage{hyperref}

\usepackage[pscoord]{eso-pic}
\newcommand{\placetextbox}[3]{
  \setbox0=\hbox{#3}
  \AddToShipoutPictureFG*{
    \put(\LenToUnit{#1\paperwidth},\LenToUnit{#2\paperheight}){\vtop{{\null}\makebox[0pt][c]{#3}}}%
  }%
}%

\makeatletter
\def\makeheaderandfooter#1{%
\def\ps@headings{%
\def\@oddhead{#1 \hbox{}\@IEEEheaderstyle\leftmark\hfil\thepage}\relax%
\def\@evenhead{#1 \@IEEEheaderstyle\thepage\hfil\leftmark\hbox{}}\relax%
}%
\def\ps@IEEEtitlepagestyle{%
\def\@oddhead{#1 \hbox{}\@IEEEheaderstyle\leftmark\hfil\thepage}\relax%
\def\@evenhead{#1 \@IEEEheaderstyle\thepage\hfil\leftmark\hbox{}}\relax%
}%
\ps@headings%
}

\makeheaderandfooter{%
\placetextbox{0.5}{1.002}{\scriptsize{%
\parbox{15cm}{\centering
This article has been accepted for publication in IEEE Transactions on Intelligent Transportation Systems. This is the author's pre-print version which has not been fully edited and content may change prior to final publication. Citation information: DOI 10.1109/TITS.2022.3214079}}}%
\placetextbox{0.58}{0.03}{\scriptsize{%
\parbox{20cm}{\raggedright
© 2022 IEEE. Personal use is permitted, but republication/redistribution requires IEEE permission. See https://www.ieee.org/publications/rights/index.html for more information.}}}%
}
 
\begin{document}
\title{Dynamic Conditional Imitation Learning for Autonomous Driving}

\author{Hesham~M.~Eraqi$^{1,2}$, 
        Mohamed~N.~Moustafa$^{1}$,
        and~Jens~Honer$^{2}$
\thanks{$^{1}$Hesham M. Eraqi and Mohamed N. Moustafa are with the Computer Science and Engineering Department, The American University in Cairo, Egypt, e-mails: {\tt heraqi@aucegypt.edu}, {\tt m.moustafa@aucegypt.edu}.}%
\thanks{$^{2}$Jens Honer and Hesham M. Eraqi are with Driving Assistance department (Valeo Schalter und Sensoren GmbH, Germany and Valeo Egypt respectively), emails: {\tt jens.honer@valeo.com}, {\tt hesham.eraqi@valeo.com}.}}%


\markboth{IEEE Transactions on Intelligent Transportation Systems}%
{IEEE Transactions on Intelligent Transportation Systems}

\maketitle

\begin{abstract}
Conditional imitation learning (CIL) trains deep neural networks, in an end-to-end manner, to mimic human driving. This approach has demonstrated suitable vehicle control when following roads, avoiding obstacles, or taking specific turns at intersections to reach a destination. Unfortunately, performance dramatically decreases when deployed to unseen environments and is inconsistent against varying weather conditions. Most importantly, the current CIL fails to avoid static road blockages. In this work, we propose a solution to those deficiencies. First, we fuse the laser scanner with the regular camera streams, at the features level, to overcome the generalization and consistency challenges. Second, we introduce a new efficient Occupancy Grid Mapping (OGM) method along with new algorithms for road blockages avoidance and global route planning. Consequently, our proposed method dynamically detects partial and full road blockages, and guides the controlled vehicle to another route to reach the destination. Following the original CIL work, we demonstrated the effectiveness of our proposal on CARLA simulator urban driving benchmark. Our experiments showed that our model improved consistency against weather conditions by four times and autonomous driving success rate generalization by 52\%. Furthermore, our global route planner improved the driving success rate by 37\%. Our proposed road blockages avoidance algorithm improved the driving success rate by 27\%. Finally, the average kilometers traveled before a collision with a static object increased by 1.5 times. The main source code can be reached at \href{https://heshameraqi.github.io/dynamic_cil_autonomous_driving}{this web page: https://heshameraqi.github.io/dynamic\_cil\_autonomous\_driving}.
\end{abstract}


\IEEEpeerreviewmaketitle

\section{Introduction}

\IEEEPARstart{D}{espite} the recent advances to achieve the promising vision of autonomous driving in terms of significantly reducing accidents \cite{who2018} and congestion \cite{brown2013autonomous}, while being environmentally and economically beneficial \cite{eraqi2019driver}, it is safe to believe that fully autonomous navigation in complex environments is still decades away \cite{janai2017computer} \cite{carla2019challenge} \cite{eraqi2021autonomous}. Autonomous vehicles employ a “sense-plan-act” design which is the basis of many robotic systems. Advanced forms of LiDAR (laser scanner, an acronym of Light Detection And Ranging), radar, and inertial measurement allowed for a more accurate and quicker sensing of the environment and surrounding objects. Nevertheless, many open challenges remain yet to be fully solved in: 1) planning the vehicle’s actions based on understating the driving scene and the interaction between its elements given the sensed data and 2) eventually commanding the vehicle’s control system steering, throttle, and brakes. The algorithmic pipeline includes tasks such as mapping, localization, driving scene perception, motion planning, and trajectory optimization which are full of open challenges \cite{fridman2019advanced} including requiring expensive data annotation, relying on heuristics and handcrafted rule-based modules, and the potential of adding unnecessary complexity to the problem. Therefore, researchers turned to train end-to-end deep neural networks to directly learn the mapping from front-facing camera data stream to driving commands.

\begin{figure}[!t]
  \centering
  \begin{minipage}{1.0\linewidth}
        \centering
        \subfigure[]{\label{fig:road_blockages_samples:sample1}
        \includegraphics[width=1.0\linewidth]{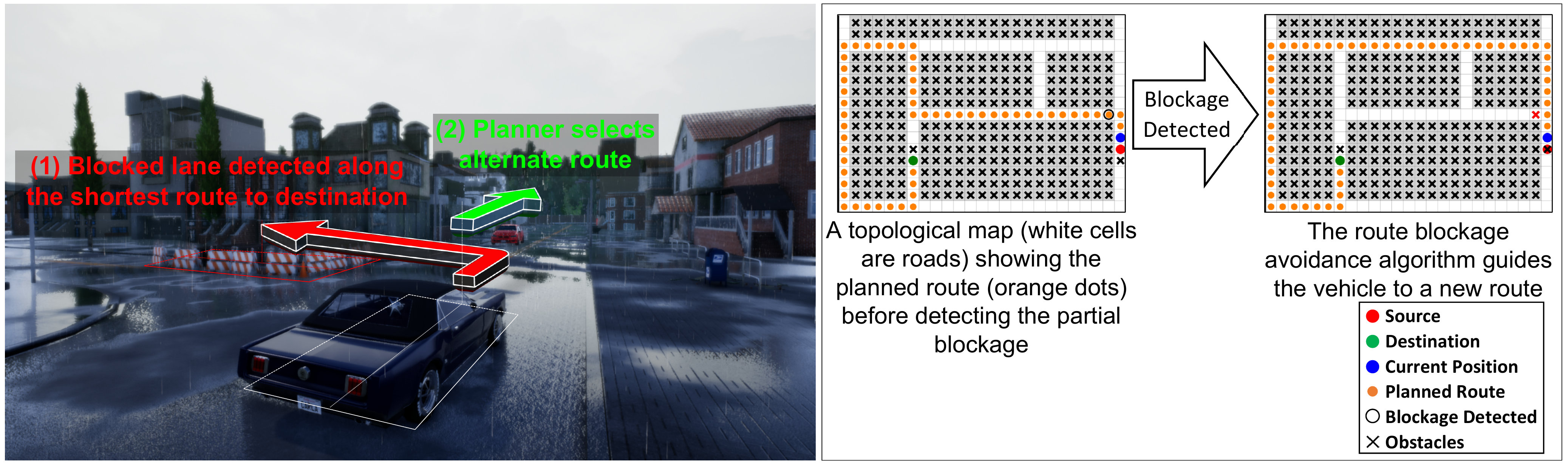}}
  \end{minipage}
  
  \begin{minipage}{1.0\linewidth}
        \centering
        \subfigure[]{\label{fig:road_blockages_samples:sample2}
        \includegraphics[width=0.5\linewidth]{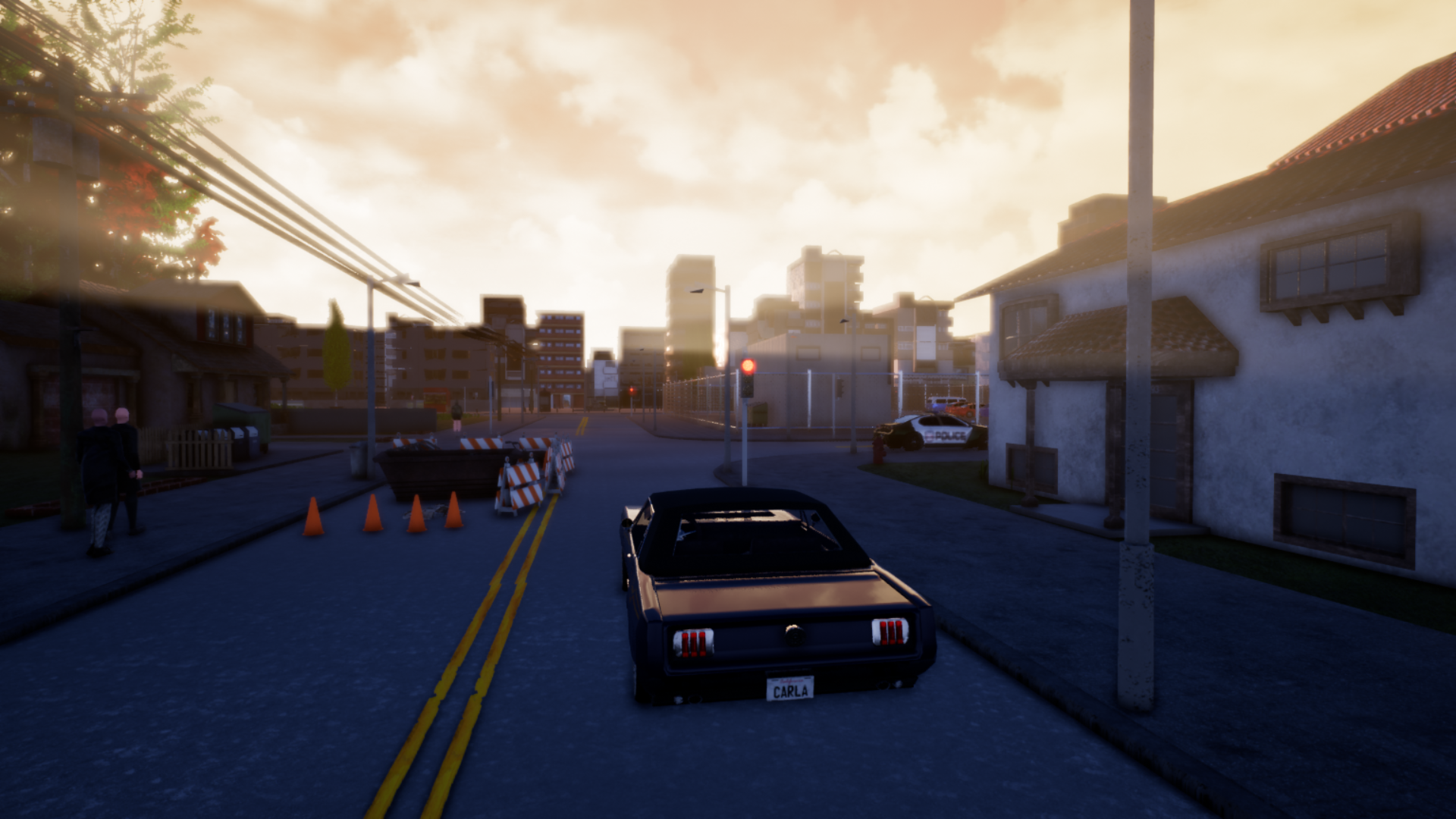}}
  \end{minipage}
  \caption{Partial road blockages added to CARLA simulator, the road blockages avoidance algorithm detects them and guides the vehicle to another route to reach the destination as demonstrated in (A), while in (B) the partial blockage is in another lane.}
  \label{fig:road_blockages_samples}
\end{figure}

Recently, there has been an increasing amount of literature adopting the end-to-end approach \cite{bojarski2016end} \cite{SteeringControl_2017_NIPS}. Such systems are demonstrated suitable when following roads and avoiding obstacles. The conditional imitation learning (CIL) method \cite{codevilla2018end} upgraded the end-to-end approach by allowing the vehicle to be automatically guided at test time to take a specific turn at an upcoming intersection to reach the destination. The model conditions imitation learning on a high-level navigational command input received from a global route planner, just as mapping applications, that instructs the model to control the vehicle to take a specific turn, go straight through an intersection, or follow lane. In \cite{dosovitskiy2017carla}, CARLA urban driving benchmark is adopted to benchmark the CIL model \cite{codevilla2018end} against other approaches to autonomous driving. CARLA \cite{dosovitskiy2017carla} is a widely used open-source simulator for autonomous car development focused on creating realistic virtual environments for the automotive industry. Many contributors constantly improve it, which makes it a comprehensive tool for simulating real-world scenarios. The benchmark results demonstrated that the CIL model is responsive to the high-level navigational commands and drives efficiently when tested in the same training environments. However, performance is found inconsistent against varying weather conditions and significantly decreases when tested in environments that are unseen during the model training. Generalization from one town's road layout and environment domains (represented by e.g. different texture sets) to other towns is a concern. Most importantly, the CIL method cannot avoid road unexpected temporary stationary blockages, as work zones in figure \ref{fig:road_blockages_samples}, which are ever-increasing in number on world roads \cite{yang2015work}. Such road blockages should be detected and the global route planner should dynamically estimate and guide the vehicle to a new route towards the destination accordingly. The aim of this work is to address two main issues of the CIL method: 1) lack of generalization to new environments and inconsistency against varying weather conditions and 2) failure to avoid static road blockages.

The contribution of this paper is two-fold. First, we provide a novel architecture that aims to tackle the CIL model \cite{codevilla2018end} challenges of lack of generalization and inconsistency against varying weathers, by extending it via fusing a LiDAR sensor input with the camera. Camera and LiDAR are primary sensor modalities for autonomous driving to capture environment semantic and geometric information respectively. The strengths of each sensor can compensate for the weaknesses of the other. The accurate LiDAR range information resolves the camera depth perception ambiguity, while the camera’s dense angular resolution compensates for LiDAR sparsity. LiDAR also is less sensitive to ambient lighting \cite{jokela2019testing}. The proposed model in this work aims to tackle the CIL model \cite{codevilla2018end} challenges of lack of generalization and inconsistency against varying weathers, by extending it via fusing a LiDAR sensor input with the camera. On CARLA simulator urban driving benchmark \cite{dosovitskiy2017carla}, the proposed model improved the autonomous driving success rate in towns unseen during the training by 52\% and improved weather consistency by 3.9 times. It outperforms the CIL model in all the different combinations of tasks and environmental setups while being trained on driving traces recorded automatically.

Our second contribution is a new efficient Occupancy Grid Mapping (OGM) method that is inspired by a part of our patent in \cite{eraqi2015resourcesaving} and used in new road blockages avoidance and topological global route planning algorithms. Detailed knowledge about the environment is useful for autonomous driving. A versatile approach to this task is to use OGM \cite{elfes1989using} to generate a map from noisy and uncertain sensor measurements. The road blockages avoidance and route planning algorithms allow the vehicle to detect unexpected road closures and to dynamically estimate a new shortest route accordingly in order to reach the destination while avoiding closed lanes or roads. Additionally, the OGM is used to rectify the proposed model output to reduce the chances of driving on the sidewalk. On CARLA benchmark \cite{dosovitskiy2017carla}, the proposed global route planner method improved the driving success rate by 37\% by providing more accurate navigational commands. The CARLA simulator and benchmark are upgraded to support testing navigation while having unexpected temporary stationary road blockages, and our road blockages avoidance algorithm improved the driving success rate by 27\% and reduced infractions with static objects by 1.5 times.

\section{Related Work}

The mediated perception approaches \cite{janai2017computer} \cite{dosovitskiy2017carla} for autonomous driving involve sub-components for detecting driving-relevant objects \cite{geiger2013vision} (as cars \cite{geiger2012we}, pedestrians \cite{li2019illumination}, lanes’ markings \cite{jung2019efficient}, or more objects), tracking of driving scene objects \cite{chang2019argoverse}, motion planning \cite{ichter2018learning}, drivable free space detection \cite{montemerlo2008junior} \cite{FreeSpaceDetection_2017}, collision avoidance \cite{CollisionAvoidance_ecta16}, mapping \cite{wirges2018object}, and more models. Then the results from these sub-components are then combined in a rule-based module that produces vehicle driving actions \cite{dosovitskiy2017carla}. Such mediated perception approach relies on scene understanding \cite{geiger2013vision} on a level that might add redundant information and unnecessary complexity; a small portion of the detected objects are relevant to driving decisions. It also requires robust solutions to open challenges in scene understanding and expensive data annotation \cite{geiger2013vision}. Direct perception \cite{chen2015deepdriving} is another approach that learns a mapping from the input camera image to several meaningful affordance indicators of the driving situation, then a rule-based controller translates them into driving actions. Those indicators are chosen via heuristics and the controller design is as expensive as the case with the mediated perception rule-based module \cite{xu2017end}. Another approach is to learn driving trajectory planning end-to-end while leaving the vehicle control component outside the end-to-end trained module \cite{bansal2018chauffeurnet} \cite{cai2019vision}.

As an alternative to the mediated perception approach, the end-to-end imitation learning approach directly maps input sensory data to driving actions via deep machine learning regression \cite{bojarski2016end} \cite{SteeringControl_2017_NIPS} \cite{janai2017computer} \cite{bojarski2017explaining}. Such behavior reflex approach optimizes the aforementioned autonomous driving sub-components simultaneously, the system self-optimizes aiming to maximize overall system performance. Unlike in the mediated perception approach, optimizing human-selected intermediate criteria doesn't guarantee maximizing overall system performance, because such criteria are selected to ease human interpretation \cite{bojarski2016end}. The major drawback of the end-to-end learning approach is that the vehicle cannot be guided to take a specific turn at an upcoming intersection. The CIL model \cite{dosovitskiy2017carla} \cite{codevilla2018end} overcomes the end-to-end approach such a limitation by training, on top of a perception Convolutional Neural Network (CNN), multiple different command-conditional modules ("branches") predicting driving commands for each possible navigational command. A topological global route planner is used to estimate a sparse list of waypoints towards the destination and based on it, the navigational command is determined for each waypoint. Based on the navigational command received from the global route planner, the proper branch is selected to control the vehicle in order to reach the destination. In \cite{montemerlo2008junior}, a hybrid A* search algorithm \cite{dolgov2008practical} is developed to predict a detailed vehicle drivable trajectory to reach the destination which is more computationally expensive than the adopted planning method in this work that estimates a sparse list of waypoints. Similarly, another approach for global planning is based on deep learning \cite{zhou2019towards} which is considerably more computationally expensive. Another weakness in such a learning-based planner is that it depends on the structure of a particular environment where the model is trained on and requires a semantic map of the environment. To cope with potential changes in certain regions in the environment, a local learning-based planner can be added which further increases the computational demands.

\begin{figure*}[!t]
  \centering
  \includegraphics[width=0.8\textwidth]{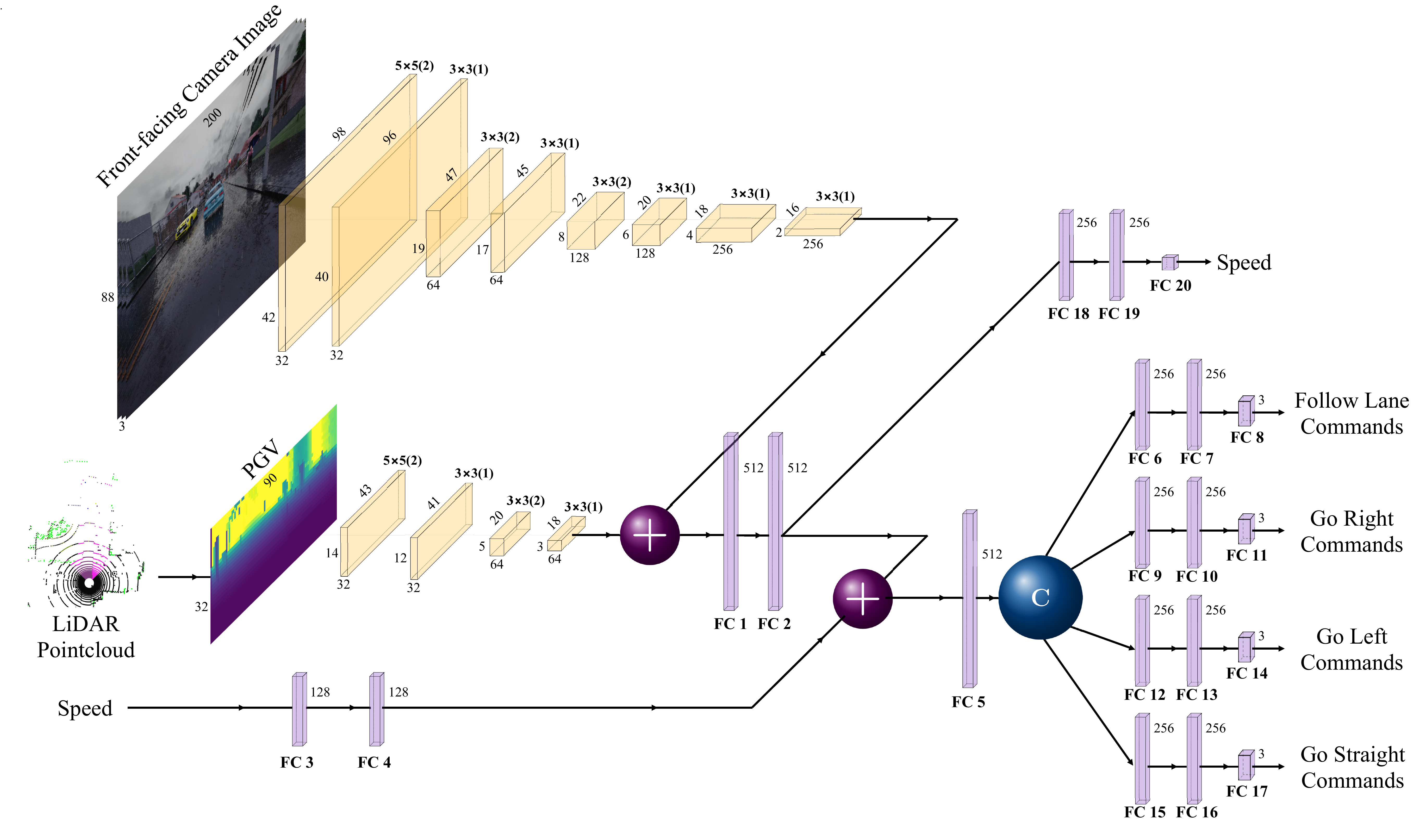}
  \caption{Proposed Network Architecture}
  \label{fig:our_model}
\end{figure*}

In \cite{dosovitskiy2017carla} \cite{codevilla2018end}, the CIL model is demonstrated to drive efficiently when deployed in the same training environments. However, performance dramatically decreases when deployed to new environments that are unseen during training time and is inconsistent against varying weathers. In addition, the CIL method cannot dynamically detect and handle road unexpected temporary stationary blockages like for example due to work zones. Various reasons have led to an ever-increasing number of road work zones of different looks and shapes even during demanding traffic levels due to maintenance, construction, and rehabilitation activities \cite{yang2015work}. Autonomous vehicles should dynamically detect such road blockages and the global route planner should estimate a new route to the destination as a result. In Junior car \cite{montemerlo2008junior}, road blockages are avoided using a drivable free space detection rule-based method. The LiDAR point cloud is analyzed against predefined thresholds and a hybrid A* planner was introduced to predict the vehicle trajectory, which makes such a method not directly compatible with the approach of learning the driving policy end-to-end. Moreover, three LiDAR sensors were used to predict a reliable vehicle trajectory. Table \ref{table:AD_approaches} summarizes the comparison between mediated perception and end-to-end approaches.

\begin{table}[!h]
    \centering
 	\setlength{\tabcolsep}{0.3em} 
    {
    \begin{tabular}{| l || c | c |} 
        \hline
        \makecell{Limitation} & \makecell{Mediated\\Perception} & \makecell{End-to-\\end} \\ 
        \hline\hline

        \makecell[l]{Requires robust solutions to open problems in\\scene understanding} & Yes & \textbf{No} \\ \hline
        
        Requires rule-based controller & Yes & \textbf{No} \\ \hline
        
        Heuristics involved & Usually & \textbf{No} \\ \hline

        \makecell[l]{Adds useless complexity in terms of detected that\\are irrelevant to driving decisions} & Yes & \textbf{No} \\ \hline
        
        \makecell[l]{Machine learning algorithm confusion caused by\\similar inputs being associated with different labels} & \textbf{No} & Yes\textsuperscript{*} \\ \hline
        
        Requires expensive training data annotation & Yes & \textbf{No} \\ \hline
        
        Can’t see the bigger picture of the driving situation & \textbf{No} & Yes\textsuperscript{*} \\ \hline
    \end{tabular}}

    \begin{tablenotes}
      \item[]\textsuperscript{*} Limitations tackled by the CIL approach
    \end{tablenotes}

\caption{The advantages and disadvantages of the mediated perception and end-to-end approaches for autonomous driving}
\label{table:AD_approaches}
\end{table}

The Occupancy Grid Mapping (OGM) algorithm is introduced in \cite{elfes1989using}. The produced map is represented in a top-view grayscale image format, where pixel intensities represent the probability of occupancy given LiDAR point cloud. Within each grid cell, the occupancy is estimated recursively using a binary Bayes filter, which creates a history effect that makes it robust against the problems of missed and false measurements. The original OGM algorithm \cite{elfes1989using} assumes map cells independence, which induces map grid cell conflicts that lead to inconsistent maps. To overcome such an inconsistency problem, a forward model \cite{thrun2001learning} is introduced to maintain dependencies between neighboring map grid cells. The algorithm is widely adopted for probabilistic localization and mapping in robotics \cite{thrun2002probabilistic}, existing autonomous vehicles \cite{montemerlo2008junior}, and vehicle trajectory prediction \cite{kim2017probabilistic}. Mapping large roadway environments with a high-resolution OGM can impose prohibitive memory requirements. Existing probabilistic quadtrees methods \cite{kraetzschmar2004probabilistic} provide compact map representations that significantly reduce the OGM memory footprint, however, they do not guarantee that the stored mapping information is utilized for the roadway information that is most important to the driving situation. This can be achieved by dynamic vehicle positioning, combined with map management techniques based on 2D ring buffers \cite{nieuwenhuisen2014obstacle} or twisted torus topology \cite{guanella2007model}. Nevertheless, vehicle dynamic positioning results into map inaccuracies due to the computationally expensive and inaccurate image sub-pixel shifting and rotation operations to compensate for ego-vehicle motion.

\section{Proposed Model}
\label{sec:model}

Figure \ref{fig:our_model} introduces our proposed network architecture. The network is end-to-end trainable, given input sensory data, the vehicle driving commands are predicted, in addition to predicted vehicle speed. The network receives the high-level navigational command $C$ as an input, alongside the image coming from a front-facing camera, LiDAR point cloud, and a measurements vector. $C$ is a turn command provided by a global route planner and acts as a switch that selects between specialized output sub-module branches. The planner should set $C$ to select the "follow lane" output branch in case of the vehicle is far away from road intersections, and during intersections, it should select the proper branch which could be to turn left, turn right, or go straight, based on the road layout and the desired destination. We adopt a topological global route planner that provides accurate navigational commands as described later in subsection \ref{subsec:route_planner}.

\begin{figure}[b] 
  \centering
  \includegraphics[width=5.0cm]{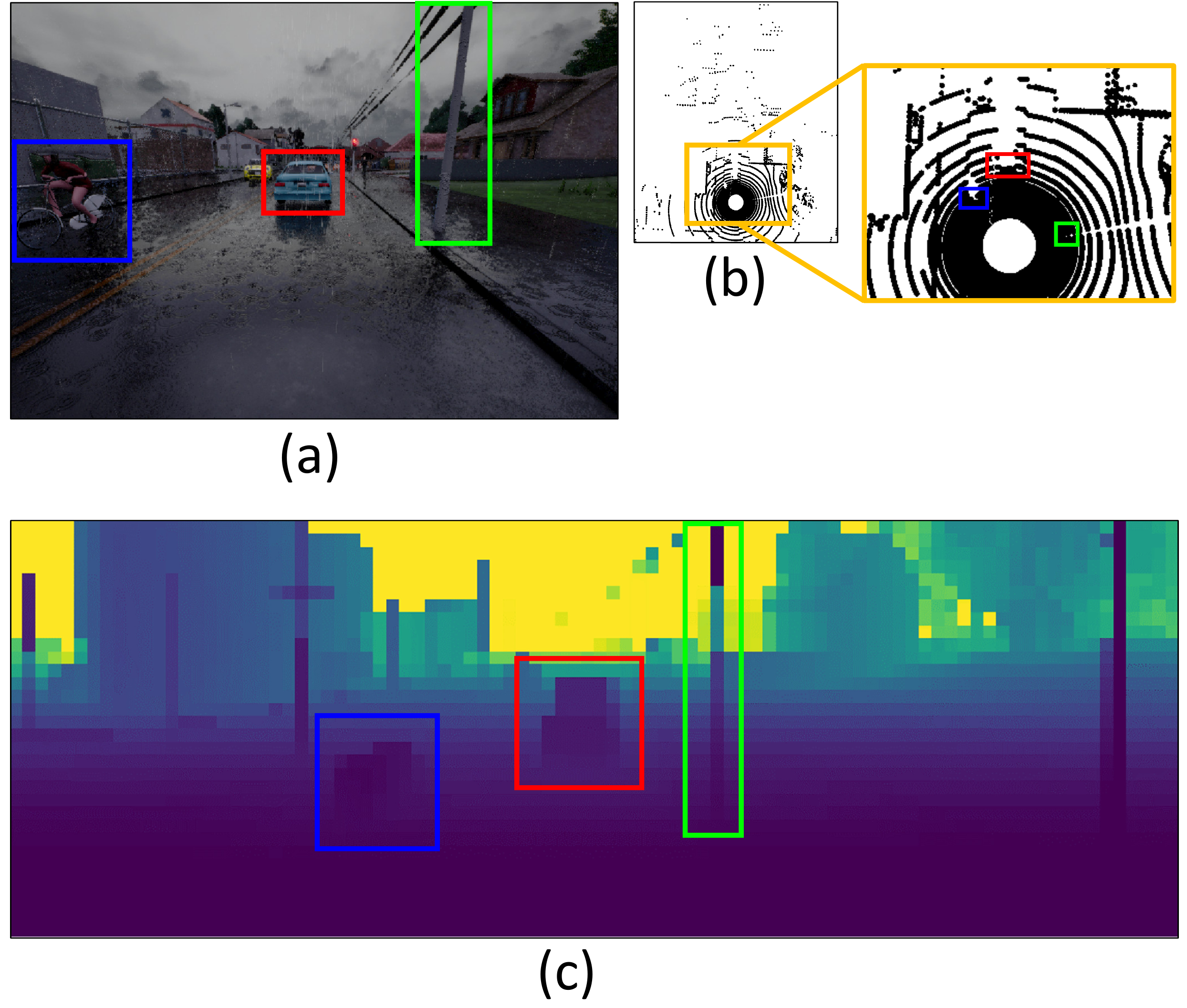}
  \caption{(a) Sample RGB camera image. (b) Corresponding LiDAR point cloud top view projection. (c) Generated PGV from the LiDAR point cloud. Three objects are matched in the figures: a vehicle, a bicyclist, and a light pole.}
  \label{fig:lidar_pgv}
\end{figure}

The camera and LiDAR input modalities are processed independently. The currently observed camera RGB image is fed into eight convolutional layers, and the associated LiDAR point cloud is encoded to a grayscale image and is then fed into four convolutional layers. The LiDAR point cloud image follows a Polar Grid View (PGV) representation. As in figure \ref{fig:our_model}, the currently observed LiDAR point cloud is encoded to a grayscale image using Polar Grid View representation (PGV). Figure \ref{fig:lidar_pgv} shows a sample camera RGB image, the corresponding LiDAR point cloud full scan top view projection, and the generated PGV which provides a 2D dense proximity spherical representation of the environment. Each LiDAR layer is associated with a PGV row, and each beam is associated with a single PGV column based on its horizontal angle. A PGV pixel holds the average depth values for all LiDAR beams that are associated with it. Such a projection-based method maps the 3D sparse point cloud into a 2D image representation that is more dense and compact compared to 3D LiDAR scan points. Consequently, standard 2D CNN can be leveraged to process those range images to achieve real-time performance \cite{cheng2021s3net}.

\begin{figure}[h] 
  \centering
  \includegraphics[width=7cm]{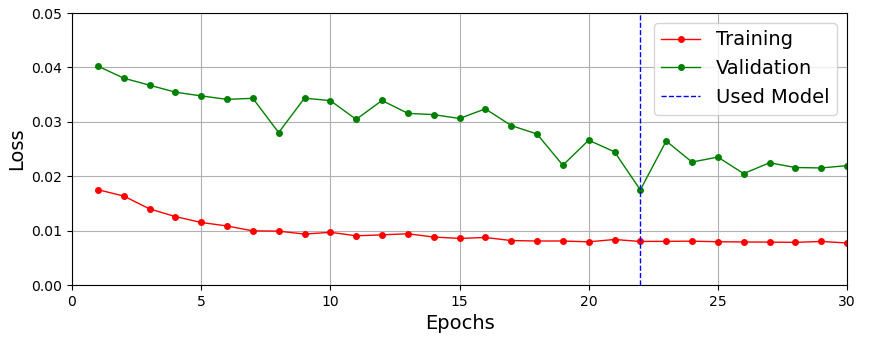}
  \caption{Training and validation losses per epoch}
  \label{fig:training_losses}
\end{figure}

We use a ReLU activation function \cite{maas2013rectifier} in all hidden layers, and a linear activation for the output layers. Figure \ref{fig:our_model} describes the number of neurons per layer, and for the convolutional layers describes kernel sizes and the used padding as well. Batch normalization is applied after all the convolutional layers, and we apply 50\% dropout after fully-connected hidden layers. For input measurements, we only use the actual vehicle speed as in \cite{codevilla2018end}. The speed, throttle, and brake values are scaled between $0$ and $1$, according to the minimum and maximum possible values. The steering wheel angle is scaled between $-1$ and $1$, with extreme values corresponding to full left and full right. The camera RGB images and the LiDAR PGV images are normalized to be in the range of $[0,1]$. For each output branch, driving actions $a$ are three-dimensional vectors that include steering wheel angle $s$, throttle $t$, and braking $b$; $a=[s,t,b]$. Given ground-truth actions $a_g$ and speeds $v_g$, and predicted actions $a$ and speeds $v$, the loss function $L$ is defined as follows: $ L = \lambda_s \left\lVert s-s_g \right\rVert^2 + \lambda_t \left\lVert t-t_g \right\rVert^2 + \lambda_b \left\lVert b-b_g \right\rVert^2 + \lambda_v \left\lVert v-v_g \right\rVert^2$, where $\lambda_s$, $\lambda_t$, $\lambda_b$, and $\lambda_v$ are empirically set to $0.5$, $0.2$, $0.15$, and $0.15$ respectively. The model is trained using Adam optimizer \cite{kingma2014adam} with $\beta_1=0.7$, $\beta_2=0.85$, and initial learning rate of 0.0002 that is multiplied by 0.5 every 10 epochs. We used mini-batches of 120 samples, where each min-batch has the same number of samples for each navigational command $C$. Half of the images in every mini-batch are augmented as described later in this section. Figure \ref{fig:training_losses} graph shows the training and validation losses per epoch. 


\begin{figure}[t]
  \centering

  \subfigure[Sample episode 1]{{\includegraphics[width=3.2cm]{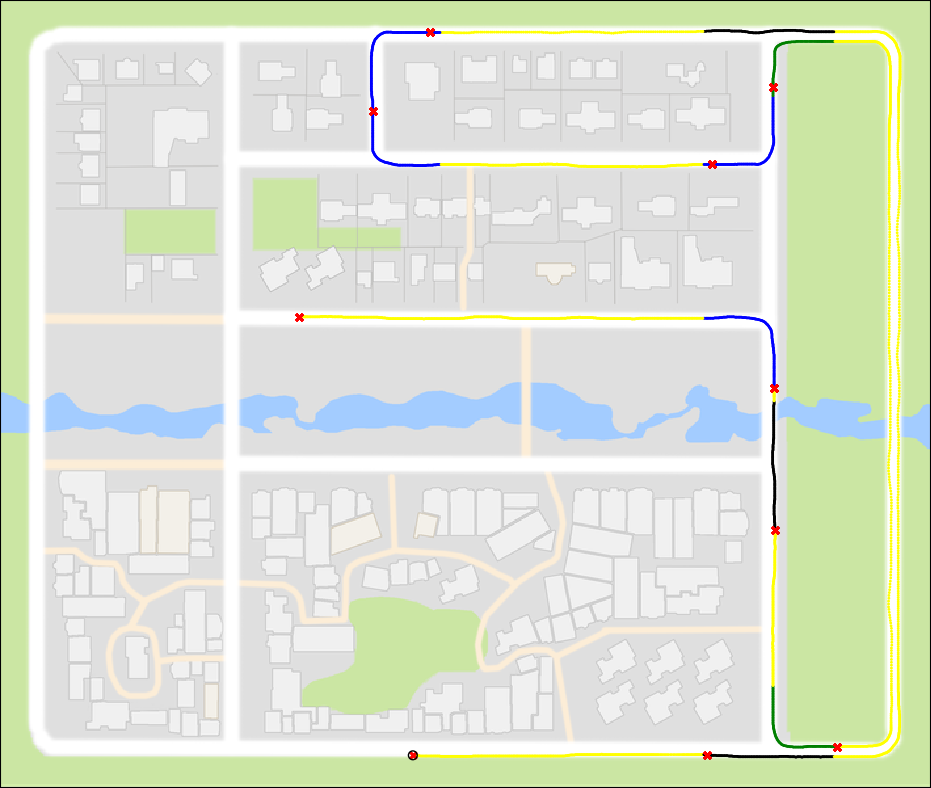} }}%
  \qquad
  \subfigure[Sample episode 2]{{\includegraphics[width=3.2cm]{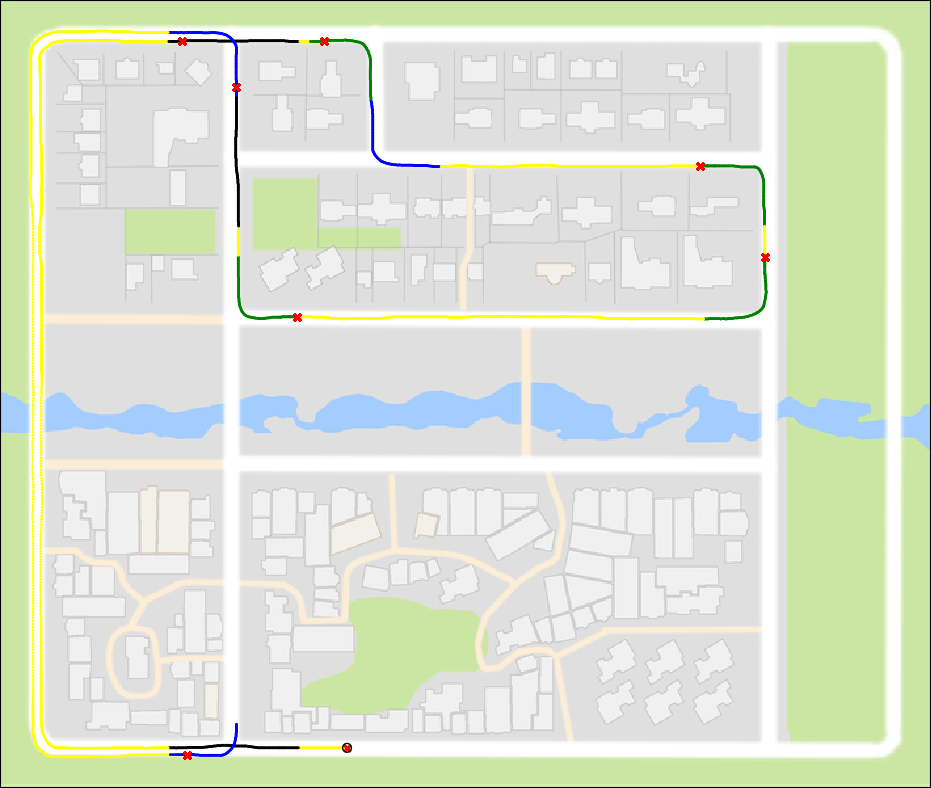} }}%

  \caption{High-level commands generated for two sample episodes during training data collection using CARLA autopilot. The red circle represents the start position and the colored trajectory shows the driving path. Colors represent the command;  yellow for "follow-lane", blue for "go left", green for "go right", and black for "go straight". Red crosses represent samples dropped out due to long traffic stopping.}
  \label{fig:dataset_high_lvl_cmd}
\end{figure}

The original CIL model is trained on a dataset collected by a human driver using CARLA simulator, who uses a signal to record his intent when approaching intersections \cite{codevilla2018end}. That signal was used as the ground-truth navigational high-level command. In contrast, our model is trained on data that is automatically recorded using two different methods. The first data collection method relies on CARLA simulator autopilot feature. In each data collection episode, the weather, traffic and pedestrians density, and vehicle starting position are randomly chosen. The ego-vehicle purposelessly follows lane and take random turning decisions in intersections and avoid obstacles for a predefined time of 10 minutes. After each episode, the navigational high-level command is generated by looking-ahead in future frames to determine the turn the vehicle decided to take. Figure \ref{fig:dataset_high_lvl_cmd} shows the generated high-level command generated for two sample episodes. In the second data collection method, we utilize CARLA route planner and PID (proportional integral derivative) controllers. In each episode a random pair of source and destination are chosen, then CARLA provides navigational waypoints that the ego-vehicle should follow to reach destination. The modular pipeline system introduced in \cite{codevilla2018end} is used to follow waypoints and avoid obstacles while making use of simulator privileged information. The training town has 2.9 km of drivable roads. For each data collection method, 200 episodes are conducted. Each episode has a number of vehicles and pedestrians uniformly randomly sampled from the ranges [30, 60] and [50, 100] respectively.

As in \cite{codevilla2018end}, temporally-correlated noise is injected into the steering during training. The noise simulates gradual drift away from the desired trajectory, then the vehicle is let to recover from these perturbations to provide the network with examples of recovery from unexpected disturbances. During model training, online data augmentation is applied to half of the mini-batch images before feeding them to the network. To augment an image, it is passed through a pipeline of a sequential series of augmentation methods. Each augmentation method in the pipeline has a predefined probability of occurrence which defines the percentage of augmented images having that method existing in their augmentation pipeline. In addition, each augmentation method has stochastic parameters to let each image be augmented differently. As an example, when adding Gaussian noise, for each image to be augmented, the Gaussian noise variance is sampled from a parameterized uniform probability distribution. Two different types of data augmentation methods are adopted. The first type is for photometric transformations: changing brightness, lighting conditions, and applying additive white Gaussian noise and Gaussian blurring \cite{imgaug}. The second type is for geometric transformations: horizontal flipping. In the case of horizontal flipping, the sign of the ground-truth steering wheel angle is flipped as well.

The proposed model is able to drive on two-lane roads (one traveling in one direction, and one traveling in the opposite direction) while having intersections and traffic lights. The training and testing towns of CARLA urban driving benchmark \cite{dosovitskiy2017carla} do not include multi-lane roads and roundabouts which are beyond the scope of this study. The model can be trained to drive on multi-lane roads as the LiDAR 360-degree field of view can enable learning lane change maneuvering; however, the adopted route planner described in subsection \ref{subsec:route_planner} needs to be adapted to support modeling more complex road networks based on schemes as Lanelets \cite{bender2014lanelets} \cite{poggenhans2018lanelet2} and OpenDrive \cite{dupuis2010opendrive}. The proposed model is also compatible to learn to drive through roundabouts as long as the route planner timely provides the navigational commands required to exit them.

\section{Efficient Occupancy Grid Mapping (OGM)}
\label{sec:ogm_and_blockage_avoidance}

\begin{figure}[h]
  \centering
  \includegraphics[width=0.28\textwidth]{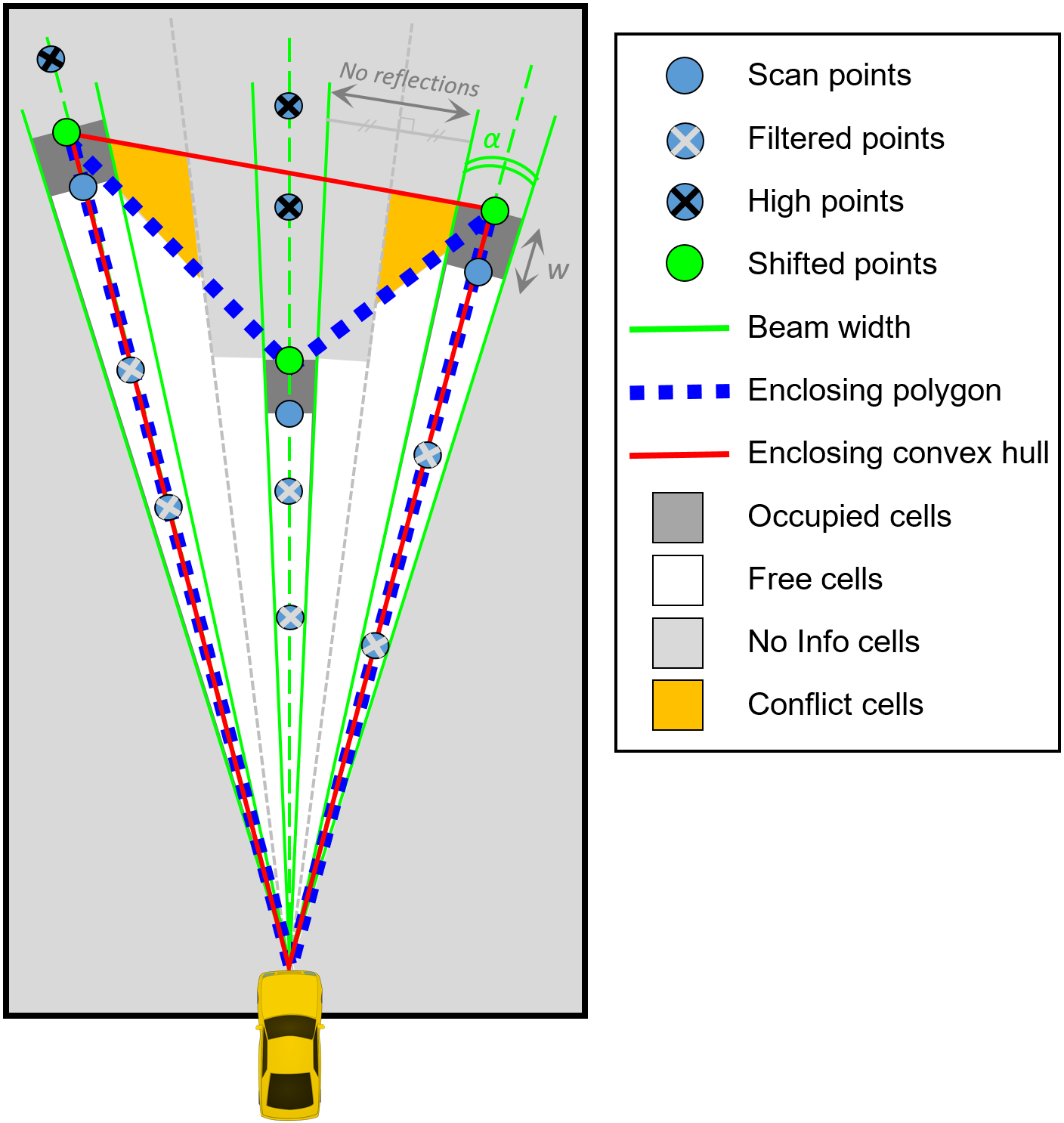}
  \caption{Comparison between considering the full scan affected area as the scan points convex hull or bounding polygon. The areas in orange color are considered "free" or as "no info" in the case of the convex hull and the polygon-based methods respectively.}
  \label{fig:ogm_area_methods}
\end{figure}

The original OGM algorithm \cite{elfes1989using} assumes map cells independence, which induces map grid cell conflicts, because a single sensor measurement may update several grid cells, which leads to inconsistent maps \cite{thrun2001learning}. To overcome such an inconsistency problem, a forward model \cite{thrun2001learning} can be used which produces more accurate OGM by maintaining dependencies between neighboring map grid cells. The forward model approach uses the Expectation-Maximization algorithm to build the map and a Laplacian approximation to model uncertainty. In this work, we introduce a new simpler and faster way to acquire OGM that preserves map quality by maintaining dependencies between neighboring map cells. As in \cite{thrun2001learning}, our inverse sensor model handles cells overlapping multiple measurements issue in \cite{elfes1989using} by generating maps from all full scan measurements at once, not incrementally on single bream measurements. Unlike \cite{thrun2001learning}, our method makes use of unreflected beams (beams with no echo returned), because they indicate important free space information. Given full scan measurements and vehicle position, the map grid cells to be updated can be determined by the convex hull or the bounding polygon of the full scan measurements. This approach assumes having dense measurements, which is a convenient assumption with LiDAR sensors. As shown in figure \ref{fig:ogm_area_methods}, both methods produce equivalent maps except for small parts around unreflected beams areas. Our method is described in Algorithm \ref{algo:ogm}. Horizontal arrow symbols in the algorithm description indicate appending to a list. The \textit{filter} function removes scan points from ground and dynamic objects based on 2D camera semantic segmentation \cite{badrinarayanan2017segnet} projected into LiDAR 3D space. The function also removes overhanging objects above a predefined height threshold (the vehicle height plus a safety buffer), like traffic signs, high tree leaves, and billboards. To make the algorithm faster, scan points could be down-sampled systematically by keeping each $k^{th}$ scan point.

\begin{algorithm}[ht]
    \DontPrintSemicolon
    \setcounter{AlgoLine}{0} 
    \label{algo:ogm}
    \caption{Proposed Occupancy Grid Map (OGM) method}
    \KwInput{$scans$: $n\times2$ array with full scan $n$ measurements, each measurement represented by its $x$ and $y$ coordinates\newline
    $P_w^t$: vehicle position in world coordinates\newline
    $P_w^{t-1}$: previous vehicle position in world coordinates\newline
    $s$: vehicle speed\newline
    $M$: OGM to be updated}
    \KwParams{$log\_odd_{free}$: log odd for free\newline
    $log\_odd_{occ}$: log odd for occupancy\newline
    $w$: wall (obstacle) depth\newline
    $\alpha$: beam width angle}
    \KwOutput{$M$: updated OGM}
    
    \BlankLine
    
    \tcp{Transform map and calculate vehicle position and orientation in map local coordinate system (run Algorithm \ref{algo:vpc})}
    $P_{local}, M = $ \textbf{position\_circle}$(s, P_w^t, P_w^{t-1}, P_{local}, M)$ 
    
    $scans = \textbf{filter}(scans)$
    
    \BlankLine
    \BlankLine
    
    \tcp{Transform to map coordinate system and shift scan points}
    $scans = scans.
        \left[ \begin{array}{cc}
            \cos(P_{local}.yaw) & -\sin(P_{local}.yaw) \\
            \sin(P_{local}.yaw) & \cos(P_{local}.yaw) \\
        \end{array} \right]
    $\;
    \textbf{allocate} 2 empty lists $angles$\ and $distances$\;
    \ForEach {\normalfont{scan point} $scan_i \in scans$}
    {
        $angles \gets \atantwo (scan_i.y-P_{local}.y, scan_i.x-P_{local}.x)$\;
        $scan_i.x \pluseq w.\cos(angles[i])$\;
        $scan_i.y \pluseq w.\sin(angles[i])$\;
        $distances \gets \sqrt{{(scan_i.x-P_{local}.x)}^2+{(scan_i.y-P_{local}.y)}^2}$\;
    }
    
    \BlankLine
    
    $area = \textbf{get\_affected\_area}(M, scans)$ \tcp{convex hull or polygon}
    \label{algo_step:affected_area}
    
    \ForEach {\normalfont{cell} $cell \in area$}
    {
        $cell_{dist} = \sqrt{{(cell.x-P_{local}.x)}^2+{(cell.y-P_{local}.y)}^2}$\;
        $cell_{angle} = \atantwo (cell.y-P_{local}.y, cell.x-P_{local}.x)$\;
        $near\_beams = $ \normalfont{list of} $i$ \normalfont{where} $|cell_{angle} - angles[i]| < \frac{\alpha}{2}$\;

        \uIf{$near\_beams$ \normalfont{list is not empty}}
        {
            $scan_{dist} = \min (distances[near\_beams])$\;
            \uIf{$cell_{dist} < scan_{dist}-w$}
            {
                $M[c] \minuseq log\_odd_{free}$\;
            }
            \uElseIf{$cell_{dist} \leq scan_{dist}$}
            {
                $M[c] \pluseq log\_odd_{occ}$\;
            }
        }
        \Else
        {
            $nearest\_beam = \argmin{i} (|cell_a - angles[i]|)$\;
            \uIf{$cell_{dist} < distances[nearest\_beam]$}
            {
                $M[c] \minuseq log\_odd_{free}$\;
            }
        }
    }
\end{algorithm}

\begin{table}[!h] 
	\def\arraystretch{1.5}
	\centering
	\caption{Occupancy Grid Mapping algorithms}
	\label{table:ogm_comparision}
	\setlength{\tabcolsep}{0.4em} 
    \renewcommand{\arraystretch}{1.2}
	\begin{tabular}{C{4.2cm}||C{1.2cm}|C{1.2cm}|C{0.6cm}}
		\hline
		Criteria & Elfes \cite{elfes1989using} & Thrun \cite{thrun2001learning} & Ours \\ \hline \hline

		Inverse sensor model accuracy & Low & \textbf{High} & \textbf{High} \\ \hline
		Incremental composition speed & Slow & \textbf{Fast} & \textbf{Fast} \\ \hline
		Uses unreflected beams information & No & No & \textbf{Yes} \\ \hline
		Memory utilization & Low & Low & \textbf{High} \\ \hline
        Map transformation speed & Low & Low & \textbf{High} \\ \hline
	\end{tabular}
\end{table}

Table \ref{table:ogm_comparision} compares our OGM method with \cite{thrun2001learning} and \cite{elfes1989using}. As in \cite{thrun2001learning}, our model adopts Bayes filtering in log-odds representation of the occupancy probabilities incremental composition, which is more computationally efficient than \cite{elfes1989using}. In the Algorithm \ref{algo:ogm}, the \textit{position\_circle} function executes our vehicle positioning method (Algorithm \ref{algo:vpc}) which aims to make the map incremental composition faster and more accurate as detailed in subsection \ref{subsec:ogm_vechile_positioning}. That method provides better memory utilization via efficient vehicle positioning on the map while preserving map accuracy by preventing map rotation and sub-pixel shifting transformations which introduce cumulative artifacts to the map.

\subsection{OGM Vehicle Positioning}
\label{subsec:ogm_vechile_positioning}

Using a global grid map is convenient for robotics applications in a controlled area \cite{grisettiyz2005improving}. But for a high-speed driving vehicle, it is not a convenient option due to limited memory and the irrelevance of old locations. Regardless of how big the memory storage is, eventually the vehicle will leave the map boundaries. Hence, a common approach is to use a local map that moves with the ego-vehicle. However, this introduces major computational burdens, namely rotation and sub-pixel shifting of the grids to compensate for ego-vehicle motion. Sub-pixel shifting is required because the vehicle motion is not necessarily a multiple of grid cell size. Additionally, both operations lead to discretization errors and create accumulated artifacts that lower the overall map quality.

\begin{figure}[!h]
  \centering
  \includegraphics[width=0.3\textwidth]{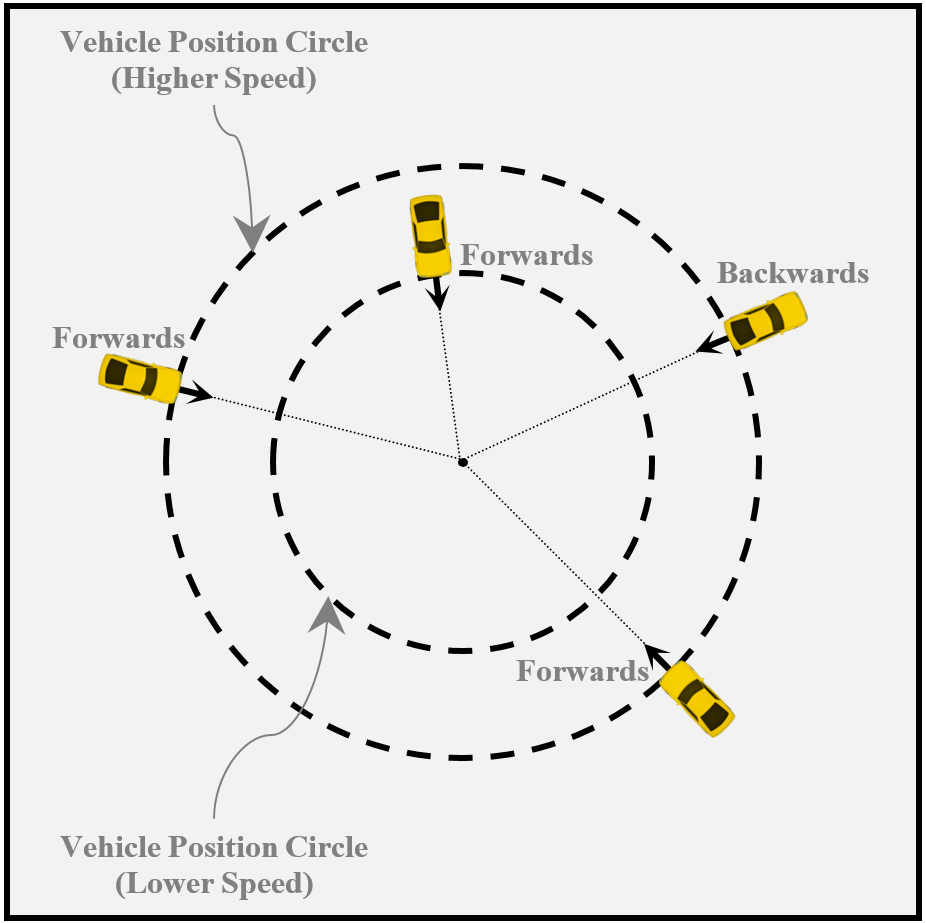}
  \caption{Position vehicle within OGM on a circle}
  \label{fig:vpc}
\end{figure}

In this work, we introduce a new method of vehicle positioning, inspired by a part of our patent in \cite{eraqi2015resourcesaving}, that allows for more accurate and computationally efficient OGM by avoiding rotation and sub-pixel shifting operations. Map sub-pixel shifting is avoided by moving the ego-vehicle within the map by the non-integer part of the required shifting, while rotation is avoided by keeping the map orientation fixed to some global coordinate system and rotating the ego-vehicle itself. The latter approach requires a square map. Yet in high-speed scenarios such as driving on a highway, the autonomous driving function is more interested in the environment in front of the vehicle. In turn, the square map reserves a lot of memory for regions of low interest if the vehicle is centered in the map. Our solution to this limitation is to grant even more freedom to the location of the ego-vehicle as shown in figure \ref{fig:vpc}. The key idea is to locate the sensor vehicle on a circle with its orientation orthogonal to the circle tangent. The circle center itself is located in the square map center. The size (radius) of the circle and the angle on which the ego-vehicle is placed on it are determined by the speed and the rotation of the vehicle ($yaw$ angle) respectively. Our algorithm is described in Algorithm \ref{algo:vpc}. Initially: $P_l^0=P_c$, $P_w^0=P_w^1$, and $M$ is a matrix filled with identical values of the average of $P_{Free}$ and $P_{Occ}$ to represent no prior occupancy information. At each full scan OGM update iteration $t$, the algorithm is executed to compensate for ego-vehicle motion, and afterwards, OGM Algorithm \ref{algo:ogm} is executed. The \textit{shift\_down} and \textit{shift\_left} functions shift up and right if they received negative shift values respectively.

\begin{algorithm}[!t]
    \DontPrintSemicolon
    \setcounter{AlgoLine}{0}  
    \label{algo:vpc}
    \caption{OGM vehicle on a circle algorithm}
    \KwInput{$s^t$: vehicle speed\newline
    $P_w^t$: vehicle position in world coordinates\newline
    $P_w^{t-1}$: previous vehicle position in world coordinates\newline
    $P_l^{t-1}$: previous vehicle position in OGM\newline
    $M^{t-1}$: OGM to be updated}
    \KwOutput{$P_l^{t}$: vehicle position in OGM\newline
    $M^{t}:$ updated OGM}
    
    \BlankLine
    
    $P_c = \textbf{get\_position\_circle}(P_w^{t}.yaw, s^t)$\;
    
    $P_l^{t}.yaw = P_w^{t}.yaw$\;
    
    \tcp{Shift map by shift value integer part}
    $P_{shift} = P_w^{t}-P_w^{t-1}+P_l^{t-1}-P_c$\;
    
    $M^{t} = \textbf{shift\_down }M^{t-1} \textbf{ by } \lfloor{P_{shift}.y\rfloor}$\;
    
    $M^{t} = \textbf{shift\_left }M^{t} \textbf{ by } \lfloor{P_{shift}.x\rfloor}$\;
    
    \BlankLine
    
    \tcp{Shift local position by the shift value fractional part}
    $P_l^{t}.x = P_c.x + \ \{P_{shift}.x\}$\;
    
    $P_l^{t}.y = P_c.y + \ \{P_{shift}.y\}$\;
\end{algorithm}

Our algorithm is a utility that can be used to position the vehicle within grid maps, regardless of the algorithm used to build these maps. It’s a pure map alignment method that saves computational time by avoiding image rotation and sub-pixel shifting. At the same time, it results in more accurate grid maps by avoiding approximations resulted from such two operations. On the other hand, the algorithm allows adopting smaller grid map sizes, and hence, less memory consumption and better real-time performance. Because it utilizes a higher proportion of the local map space for important areas based on vehicle movement trajectory. Our method remains compatible with existing efficient map management techniques of representing map cells in terms of 2D ring buffers \cite{nieuwenhuisen2014obstacle} or twisted torus topology \cite{guanella2007model}.

\section{Road Blockages Avoidance}

\subsection{Global Route Planner}
\label{subsec:route_planner}

We adopt a topological global route planning algorithm that is similar to the one used in \cite{codevilla2018end} and \cite{dosovitskiy2017carla}. The implementation logic is upgraded to provide more accurate high-level navigational commands $C$ and to execute faster. As in Algorithm \ref{algo:route_planner}, a command from four possibilities of turn left, turn right, go straight, and follow lane is returned based on the vehicle destination GPS coordinates and orientation which defines the arrival lane, using the vehicle GPS and compass. As in \cite{codevilla2018end}, the planning is carried out on a one-way roads grid map for simplicity and to make planning faster; the A* search algorithm is carried out after setting the map cell after the destination and the cell behind the vehicle as occupied, i.e.; putting 'walls'. In the route planner in \cite{codevilla2018end}, the A* algorithm is executed to re-evaluate the shortest path towards the destination every time the vehicle travels to a new map cell.

\SetInd{0.3em}{0.85em} 
\begin{algorithm}[!ht]
    \DontPrintSemicolon
    \label{algo:route_planner}
    \caption{Global Route planner algorithm}
    \KwInput{$roads$: list of town roads, a road is represented by the GPS coordinates of its start and end\newline
    $car\_gps$: vehicle GPS coordinates\newline
    $car\_compass$: vehicle orientation in the world\newline
    $dest\_gps$: GPS coordinates of the desired destination\newline
    $dest\_orient$: the destination orientation (defines the desired arrival lane)\newline
    $ogm$: OGM with vehicle position in it}
    \KwParams{$res$: planning map resolution, the output command is constant within a cell\newline
    $far\_inters$: number of map cells to the nearest intersection to decide it is far away\newline
    $inter\_exited$: number of cells away from a visited intersection to decide the vehicle left it}
    \KwOutput{$C$: a high-level navigational command, used by driving model}

    \BlankLine

    \tcp{Initializing global variables}
    \If{\normalfont{algorithm called for first time}}
    {
        $world\_graph = \textbf{directed\_weighted\_graph}(roads)$ \tcp{A node for each roads intersection, weights are road distances}
        
        $intersects =$ list of nodes in $world\_graph$ with $edges > 2$
        
        $map = \textbf{grid\_map}(world\_graph, res)$ \tcp{a one-way roads map where graph nodes are connected via road free cells and all the other cells are marked as walls}
    
        $map = \textbf{add\_destination\_wall}(map, \textbf{gps\_to\_cell}(dest\_gps), \newline dest\_orient)$ \tcp{add a wall in the cell after the destination position to plan to arrive in the desired lane}
        
        $prev\_cell, next = None$ \& $route\_exited = False$
    }
    
    $car\_cell, dest\_cell = \textbf{gps\_to\_map\_cell}(car\_gps, dest\_gps)$

    \If{$car\_cell = dest\_cell$ \textbf{ AND } $car\_compass = dest\_orient$}
    {
        \textbf{return } $GOAL\_REACHED$
    }
    
    \If{$prev\_cell \neq None \textbf{ AND } car\_cell \neq prev\_cell$}
    {
        $dist = \textbf{route\_distance}(route, car\_cell, route[next])$
        
        \If{$dist \leq 1$} {
            $prev\_cell = car\_cell$ \& $next \pluseq dist$
        }
        \lElse{$route\_exited = True$ \tcp*[f]{if $car\_cell$ is not on $route$}}
    }%
    
    \tcp{Re-estimate route when needed}
    
    $map, road\_blocked = road\_blockages(map, route, ogm)$ \tcp{Run Algorithm \ref{algo:road_blockages} to detect road blockages}
    \label{algo_step:estimate_road_blockages}

    \If{$road\_blocked \textbf{ OR } route\_exited \textbf{ OR } prev\_cell = None$}
    {
        $road\_blocked, route\_exited = False$ \& $prev\_cell = car\_cell$
        
        $map = \textbf{add\_car\_wall}(map, car\_cell, car\_compass)$ \tcp{add a wall in the cell behind the car current cell}
        
        $route = \textbf{a\_star}(map, car\_cell, dest\_cell)$ \tcp{A\textsuperscript{*} from $car\_cell$ to $dest\_cell$}
        
        $next = 1$ \tcp{cell after vehicle cell index in $route$}
        
        $cmds = $ list of $route.count$ length and $'Follow\_Lane'$ values
        
        \For{$i \textbf{ from } 0 \textbf{ to } route.count$}
        {
            \If{$route[i] \textbf{ is in } intersects$}
            {
                $s = \textbf{normalized\_cross\_product}(\overrightarrow{route[i],route[i+1]}, \newline \overrightarrow{route[i-1],route[i]})$
                
                \lIf{$s < -0.1$} {$cmd = 'Go\_Right'$}
                \lElseIf{$s > 0.1$} {$cmd = 'Go\_Left'$}
                \lElse{$cmd = 'Go\_Straight'$}
                
                $cmds[i-far\_inters:i+inter\_exited] = cmd$
            }
        }
    }
    
    \For{$i \textbf{ from } 0 \textbf{ to } route.count$}
    {
        \If{$route[i] = car\_cell$}
        {
            $C = cmds[i]$ \& \textbf{break}
        }
    }
\end{algorithm}

\begin{figure}[h]
    \centering
    \centering
    \includegraphics[width=0.85\linewidth]{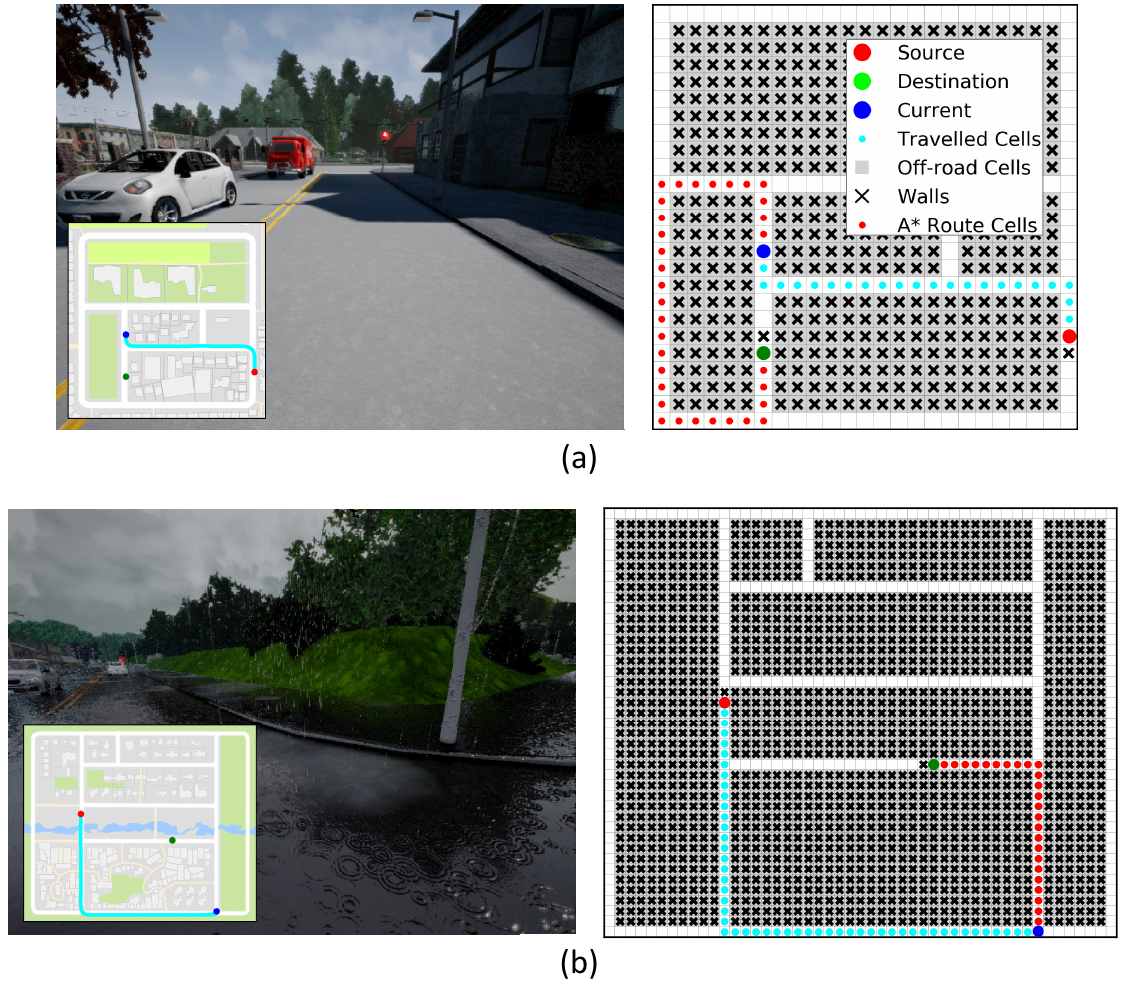}
    \caption{Two sample snapshots for the proposed global route planning during deployment. The front-facing camera image, the world map, and the planning map are shown for each snapshot. The town in sample \ref{fig:route_planner}(b) is larger in area. In sample (a) snapshot, our planner returns a "follow lane" navigational command, while the planner adopted in \cite{codevilla2018end} and \cite{dosovitskiy2017carla} returns a "go left" navigational command too early which causes the vehicle to invade the opposite lane. In sample (b), both planners return a "go left" command.}
    \label{fig:route_planner}
\end{figure}

Our planner is different from the planner in \cite{codevilla2018end} in two respects. Firstly, we make the process faster by executing the A* algorithm only when the vehicle exits the latest planned shortest path towards the destination or when detecting a road blockage along the path. This is more similar to the global route planner in \cite{montemerlo2008junior} that is carried out at each checkpoint or when facing a road blockage. Secondly, and more importantly, our algorithm provides more accurate commands around road intersections than \cite{codevilla2018end} as shown in the sample in figure \ref{fig:route_planner}(a). Figure \ref{fig:route_planner} shows two sample snapshots for the algorithm during deployment from different towns. In sample \ref{fig:route_planner}(a), our planner returns a ”follow lane” navigational command, while the planner adopted in \cite{codevilla2018end} returns a ”go left” navigational command too early. In step \ref{algo_step:estimate_road_blockages} in Algorithm \ref{algo:route_planner}, the function \textit{road\_blockages} executes Algorithm \ref{algo:road_blockages} to check if the planned road ahead has any blockages. The function adds walls to the planning map corresponding to detected road blockages. As discussed later in subsection \ref{subsec:road_blockages_avoidance}, the added walls are directed; the A* search algorithm decides whether a cell with a directed wall is blocked or free to traverse based on the direction of reaching it from previous cells.

\subsection{OGM-based Road Blockages Avoidance}
\label{subsec:road_blockages_avoidance} 

We updated CARLA simulator to support the insertion of road blockages programmatically as shown in the samples in Figure \ref{fig:road_blockages_samples}. Algorithm \ref{algo:road_blockages} describes our method to avoid road blockages based on OGM that is inspired by a part of our patent in \cite{eraqi2020DCIL}. The algorithm is called by the global route planner Algorithm \ref{algo:route_planner}. The inputs are the planning map, an ordered list of planning map cell coordinates of the route towards the destination, and the OGM along with its associated vehicle position information. The cell coordinates are projected and shifted to the lanes' center and are used as control points that are up-sampled and smoothed with a Bezier curve. The OGM is queried for occupancy by the Bezier curve points based on thresholds, if occupancy is detected along the curve, the flag $road\_blocked$ is set to $True$ and the planning map is updated by adding the blocked cell, so the global route planner calculates a new $route$ that avoids the detected road blockage and consequently provides the appropriate navigational commands to the model to avoid the blockage. In case of a false positive, the global planner will cause the vehicle to unnecessarily choose a longer route to reach the destination.

\SetInd{0.3em}{0.85em} 
\begin{algorithm}[t]
    \DontPrintSemicolon
    \label{algo:road_blockages}
    \caption{Road blockages avoidance algorithm}
    \KwInput{$planning\_map$: global route planning map\newline
    $route$: an ordered list of planner route cells\newline
    $ogm$: OGM with vehicle position in it}
    \KwParams{$route\_pts$: the number of cells in $route$ to consider\newline
    $P_{occ}$: OGM occupied cell probability threshold\newline
    $w$: the width and height of the OGM slice around a waypoint\newline
    $cell\_occupied\_pts$: minimum number of OGM occupied cells to block a planning map cell}
    \KwOutput{$planning\_map$: planning map updated with added walls if road blockages detected\newline
    $road\_blocked$: a flag set to $True$ if a road blockage is detected}

    \BlankLine

    $route = route[0:route\_pts]$ \& $road\_blocked = False$
    
    $prev\_cell = route[0]$  \tcp{route[0] is vehicle cell}

    $route\_lane = \textbf{shift\_to\_lane}(\textbf{cell\_to\_gps}(route))$ \tcp{Project to nearest point middle of the road then shift car lane lane center}
    
    $waypoints = \textbf{smooth\_with\_bezier}(route\_lane)$ \tcp{Up-sampled Bezier curve defined by $route\_lane$ control points}
    
    $waypoints\_ogm = \textbf{gps\_to\_ogm}(waypoints, ogm)$
    
    \ForEach {\normalfont{cell} $c \in route[1:end]$}
    {
        $points = $ \normalfont{list of points} \normalfont{in} $waypoints\_ogm$ \normalfont{located inside} $c$

        $occ = 0$

        \ForEach {\normalfont{point} $p \in points$}
        {
            $OGM_s =$ \normalfont{slice in} $ogm$ \normalfont{of width and height} $w$ \normalfont{around} $p$
            
            $occ \pluseq $ \normalfont{number of cells in} $OGM_s$ \normalfont{with value} $> P_{occ}$
            
            \If{$occ > cell\_occupied\_pts$}
            {
                $blockage\_direction = \textbf{get\_direction}(c, prev\_cell)$
            
                $planning\_map = \textbf{add\_directed\_wall}(planning\_map, c, blockage\_direction)$
                
                $road\_blocked = True$ \& \textbf{break}
                
            }
        }
        
        $prev\_cell = c$
    }
\end{algorithm}

\begin{table*}[!b] 
	\def\arraystretch{1.5}
	\centering
	\caption{In each town in the CARLA urban driving benchmark \cite{dosovitskiy2017carla}, there are 24 experiment sets. Each set has 25 test scenarios (1200 in total) representing a combination of a driving task and a weather condition. "S", "O", "N", and "DN" stand for "straight", "one (single) turn", "navigation", and "dynamic navigation (along moving vehicles and pedestrians) " tasks respectively.}
	\label{table:codevilla_experiments_description}
	\setlength{\tabcolsep}{0.4em} 
    \renewcommand{\arraystretch}{1.05}
    \small 
	\begin{tabular}{c||c|c|c|c|c|c|c|c|c|c|c|c|c|c|c|c|c|c|c|c|c|c|c|c}
		\hline
Experiment ID	                     & 1 & 2 & 3 & 4	                                                              & 5 & 6 & 7 & 8		                                                     & 9 & 10 & 11 & 12	                                                             & 13 & 14 & 15 & 16		                                                   & 17 & 18 & 19 & 20		                                                    & 21 & 22 & 23 & 24                                                           \\ \hline \hline
Task								 & S                  & O                  & N                  & DN		      & S                  & O                  & N                  & DN		 & S                  & O                   & N                   & DN		     & S                   & O                   &  N                  & DN		   & S                   & O                   & N                   & DN	    & S                   & O                   & N                   & DN        \\ \hline
\makecell{Weather\\Condition}  & \multicolumn{4}{c|}{\makecell{Clear Afternoon\\(Train)}}     & \multicolumn{4}{c|}{\makecell{Wet Noon\\(Train)}}	     & \multicolumn{4}{c|}{\makecell{Wet Cloudy Noon\\(Test)}}     & \multicolumn{4}{c|}{\makecell{Hard Rain Noon\\(Train)}}   & \multicolumn{4}{c|}{\makecell{Clear Sunset\\(Train)}}    & \multicolumn{4}{c}{\makecell{Soft Rain Sunset\\(Test)}}  \\ \hline
	\end{tabular}
\end{table*}

\begin{figure}[!b]
  \centering
  \includegraphics[width=9.0cm]{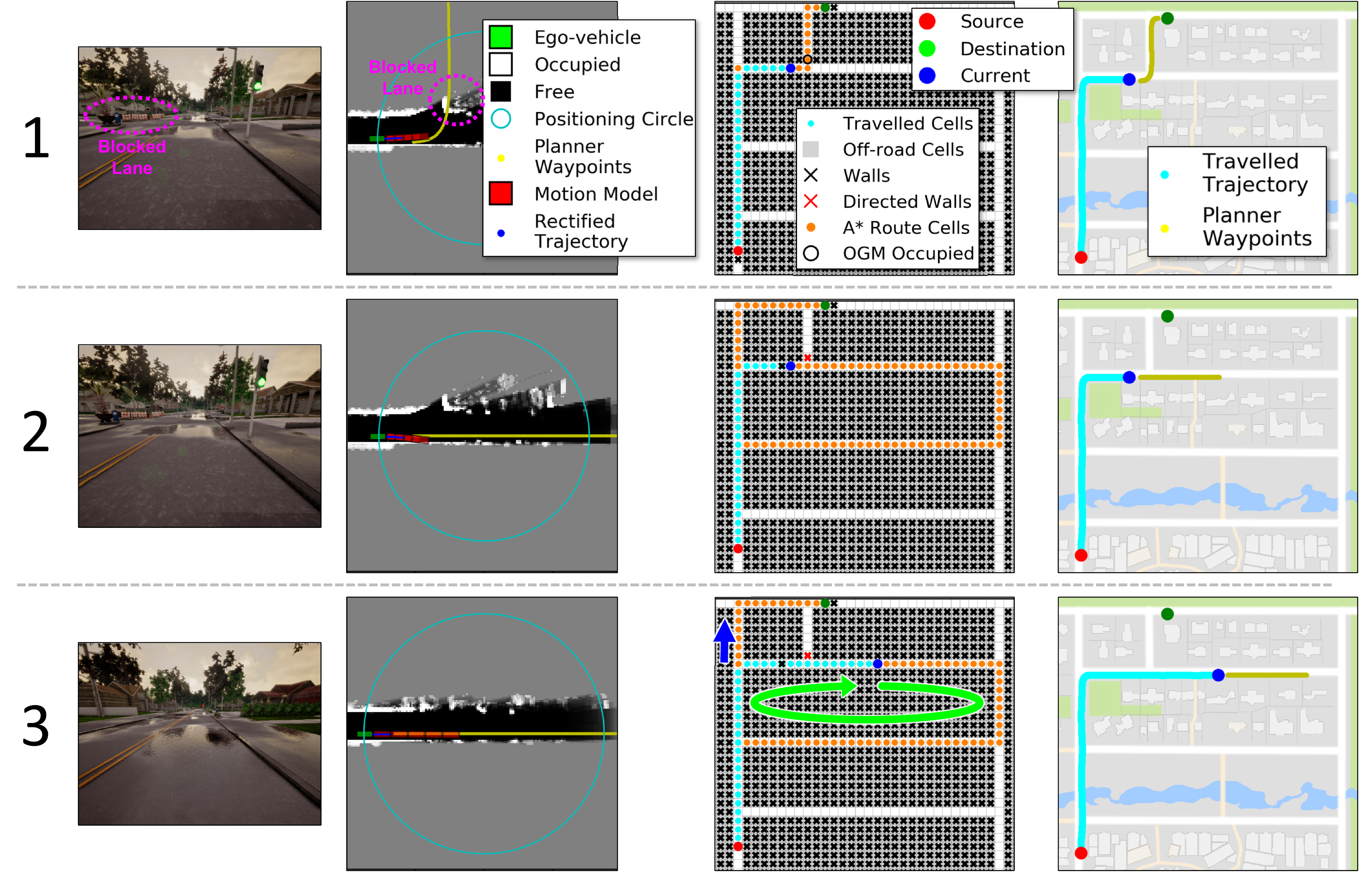}
  \caption{Three snapshots from a navigation scenario demonstrating the road blockages avoidance method. In the first snapshot, a blockage is detected, and in the second snapshot, the planner successfully re-routes the vehicle to avoid it. If the blockage added wall is directed, the vehicle continues along the route following the blue arrow in the third snapshot route planning map to reach the destination. If the added wall is removed once the vehicle leaves its area, the vehicle will be stuck in the navigation loop indicated by the green arrow in case that the shortest path towards the destination is re-evaluated.}
  \label{fig:navigation_loop}
\end{figure}

The walls added to the planning map due to detected road blockages are directed. The \textit{get\_direction} function calculates the direction from four possibilities of right, up, left, and down from the cell that should have the added wall to its preceding cell in $route$. The route planner decides whether the cell with a directed wall is blocked or free to traverse based on the direction of reaching it. The concept of directed walls enables handling partial road blockages where a single lane is occupied while the other lanes could be free to navigate while at the same time leveraging the advantages of planning on a one-way roads map. Figure \ref{fig:road_blockages_samples} shows examples for such partial blockages. The workaround of removing the added walls after the vehicle leaves its area, when blockages become irrelevant, instead of having walls directed causes the problem of having the vehicles infinitely looping the same course in some situations as described in Figure \ref{fig:navigation_loop} scenario. The A* search algorithm was set to prioritize fewer turns when having multiple shortest paths having the same Manhattan distance.

The algorithm considers only the near future waypoints through the parameter $route\_pts$, to save computational power that might be wasted beyond the information in the OGM due to the limited LiDAR range. If $route\_pts$ is set too small, the global planner will detect road blockages too late for the model to be able to avoid them. The OGM history effect makes it more accurate in near ranges from the vehicle, hence adding walls in cells farther away from the vehicle should be made less likely by making the involved thresholds function of the distance from the vehicle. In addition, based on the OGM and the ego-vehicle current position and speed, a simple rule-based method is employed which determines if the ego-vehicle should slightly reduce or increase the model predicted steering angle to avoid collision with static obstacles. Given the vehicle wheel radius and base, a kinematic bicycle model is utilized to estimate the vehicle position and orientation in near-term future frames. As shown in figure \ref{fig:navigation_loop} OGM's, the motion model estimates discrete (five in the figure example) vehicle states within three seconds in the future. The discrete trajectory is equidistantly resampled and smoothed giving dense points, and the points within a predefined short distance threshold from the vehicle are considered (named rectified trajectory in the figure). The OGM is queried for being free along the rectified trajectory points. If any points along the trajectory are not free, the steering angle is gradually increased or decreased within a small range until all the trajectory points become possibly free in the OGM. Such a steering rectification method is especially useful when the road blockages' visual elements are not well-represented in the model training data.

\begin{table*}[!t] 
	\centering
	\caption{Driving success rate average percentage, average percentage of distance to goal traveled is between parentheses}
	\label{table:success_rates_distance_to_goal_results}
	\setlength{\tabcolsep}{0.1em} 
    \renewcommand{\arraystretch}{1.0} 
	\begin{tabular}{|c|c||c|c|c|c|}
		\hline
		\multirow{3}{*}{Task}								& \multirow{3}{*}{Model}							            & \multicolumn{4}{c|}{\makecell{Percentages of average success rate and distance to goal}}	\\ \cline{3-6}
															&													            & \multicolumn{2}{c|}{Training town}	& \multicolumn{2}{c|}{New town} 					\\ \cline{3-6}
															&													            & Training weathers	    & New weathers	& Training weathers	& New weathers 	                \\ \hline \hline
\multirow{5}{*}{Straight} & Camera only, \cite{dosovitskiy2017carla} results & 95 (-) & 98 (-) & 97 (-) & 80 (-) \\ 
& Camera only, \cite{dosovitskiy2017carla} pre-trained & 99 (97.24) & \textbf{100 (100.00)} & 89 (90.38) & 92 (92.70) \\ 
& Camera only (trained on our data) & \textbf{100 (100.00)} & \textbf{100 (100.00)} & 99 (95.74) & \textbf{100 (100.00)} \\ 
& Camera only (trained on our data, no augmentation) & 93 (91.46) & 88 (87.06) & 88 (87.67) & 86 (85.71) \\ 
& Camera + LiDAR & \textbf{100 (100.00)} & \textbf{100 (100.00)} & \textbf{100 (100.00)} & \textbf{100 (100.00)} \\ 
& \textbf{Camera + LiDAR (Our Route Planner)} & \textbf{100 (100.00)} & \textbf{100 (100.00)} & \textbf{100 (100.00)} & \textbf{100 (100.00)} \\ \hline \hline
\multirow{5}{*}{\makecell{Single\\Turn}} & Camera only, \cite{dosovitskiy2017carla} results & 89 (-) & 90 (-) & 59 (-) & 48 (-) \\ 
& Camera only, \cite{dosovitskiy2017carla} pre-trained & 88 (82.71) & 94 (85.49) & 56 (54.81) & 74 (60.60) \\ 
& Camera only (trained on our data) & 97 (97.33) & 98 (97.72) & 57 (56.09) & 72 (67.18) \\ 
& Camera only (trained on our data, no augmentation) & 73 (67.29) & 72 (65.91) & 49 (43.27) & 56 (52.57) \\ 
& Camera + LiDAR & \textbf{100 (100.00)} & \textbf{100 (100.00)} & \textbf{92 (90.05)} & \textbf{92 (91.52)} \\ 
& \textbf{Camera + LiDAR (Our Route Planner)} & \textbf{100 (100.00)} & \textbf{100 (100.00)} & \textbf{92 (88.17)} & \textbf{92 (88.17)} \\ \hline \hline
\multirow{5}{*}{Navigation} & Camera only, \cite{dosovitskiy2017carla} results & 86 (-) & 84 (-) & 40 (-) & 44 (-) \\ 
& Camera only, \cite{dosovitskiy2017carla} pre-trained & 78 (88.61) & 84 (89.34) & 35 (9.68) & 58 (45.37) \\ 
& Camera only (trained on our data) & 87 (91.13) & 88 (92.46) & 33 (16.92) & 34 (16.93) \\ 
& Camera only (trained on our data, no augmentation) & 65 (64.86) & 66 (69.06) & 28 (15.02) & 25 (14.75) \\ 
& Camera + LiDAR & 92 (92.70) & 92 (92.71) & 68 (76.95) & 68 (76.86) \\ 
& \textbf{Camera + LiDAR (Our Route Planner)} & \textbf{96 (96.02)} & \textbf{96 (96.04)} & \textbf{100 (100.00)} & \textbf{100 (100.00)} \\ \hline \hline
\multirow{5}{*}{\makecell{Dynamic\\Navigation}} & Camera only, \cite{dosovitskiy2017carla} results & 83 (-) & 82 (-) & 38 (-) & 42 (-) \\ 
& Camera only, \cite{dosovitskiy2017carla} pre-trained & 80 (88.36) & 74 (81.53) & 28 (17.35) & 54 (35.13) \\ 
& Camera only (trained on our data) & 84 (91.03) & 82 (87.34) & 26 (9.53) & 30 (29.41) \\ 
& Camera only (trained on our data, no augmentation) & 58 (59.53) & 58 (61.41) & 24 (11.72) & 23 (12.07) \\ 
& Camera + LiDAR & 86 (93.02) & 86 (92.89) & 53 (37.51) & 64 (59.90) \\ 
& \textbf{Camera + LiDAR (Our Route Planner)} & \textbf{94 (98.16)} & \textbf{96 (98.59)} & \textbf{89 (80.88)} & \textbf{88 (77.80)} \\ \hline \hline
	\end{tabular}
\end{table*}

\section{Experimental Results}
\label{sec:results}

We adopt the experimental setup of the CARLA urban driving benchmark \cite{dosovitskiy2017carla} to evaluate the proposed model. The benchmark is composed of four different tasks that are carried out in two towns and six weather conditions. The test town and weather conditions are fully unseen during training. For each combination of a task, a town, and a weather set, testing is conducted over 25 different test scenarios having predefined start and destination locations, this gives a total of 1200 test scenarios for each model under test as described in table \ref{table:codevilla_experiments_description} for one town. In the benchmark, each test scenario in the 'Dynamic Navigation' (navigation in traffic) task has 50 pedestrians moving in the driving town and 20 or 15 vehicles for towns 1 and 2 respectively. The parameters for the PID controllers we used during training data recording are tuned in the training town and weathers. Town 1 and 2 have 2.9 km and 1.9 km of drivable two-lane roads in a suburban environment respectively. Both towns include 3-way intersections; 7 and 12 intersections in towns 1 and 2 respectively and do not contain roundabouts.

A test scenario is considered successful if the vehicle reaches the destination within a predetermined deadline. The deadline (maximum allowed time to reach the destination) is set to the time needed to reach the destination along the shortest route at a low speed of 10 km/h as followed in \cite{dosovitskiy2017carla} and \cite{chen2020learning}. A model driving at that low speed has low chances to succeed in a test scenario as it has not to stop due to traffic (for the ‘Dynamic Navigation’ tasks) or red lights nor to slow down to avoid collisions, while at the same time it has to estimate and follow the shortest route towards the destination without missing a single turn. In addition, such a low-speed deadline allows models that drive at higher, and more reasonable, speeds to succeed in a test scenario even if the shortest route is not followed. The proposed model average vehicle speed over all the benchmark test scenarios was 25.428 km/h compared to 18.697 km/h for the pre-trained model in \cite{codevilla2018end}. The actual vehicle speeds while moving are higher than those two numbers as the averaging includes traffic stopping moments.

\subsection{Success Rate and Distance to Destination}
\label{subsec:success_rates_distance_to_goal_results}

Table \ref{table:success_rates_distance_to_goal_results} benchmarks the proposed model before and after LiDAR fusion with the state-of-the-art CIL model \cite{codevilla2018end} on the CARLA urban driving benchmark \cite{dosovitskiy2017carla}. The table reports the autonomous driving success rate on different tasks and test conditions, and the average percentage of distance to goal traveled is available between parentheses. The latter metric provides additional insight that cannot be inferred from the success rate. It is not included in the original benchmark, thus we include the results we record from deploying the publicly available CIL pre-trained model. The table includes an ablated model trained without data augmentation.

Table \ref{table:success_rates_distance_to_goal_results} benchmark results for the pre-trained model in \cite{dosovitskiy2017carla} ("Camera only, \cite{dosovitskiy2017carla} pre-trained" model) are correlated to the results reported in \cite{dosovitskiy2017carla} ("Camera only, \cite{dosovitskiy2017carla} results" model), but are not matching exactly. These discrepancies are due to randomness in evaluation and differences in texture appearance compared to the earlier version of CARLA used in the \cite{dosovitskiy2017carla}. Additionally, there are two known sources of non-determinism: 1) textures loading time is not deterministic in the underlying game engine which leads to appearance differences, and 2) the simulator pedestrians algorithms are non-deterministic. The third model in the table ("Camera only (trained on our data)") trains the same original CIL model from scratch on our dataset. The results are correlated with the preceding two models ("Camera only, "\cite{dosovitskiy2017carla} results" and "Camera only, "\cite{dosovitskiy2017carla} pre-trained"), but are less accurate, especially in the harder tasks and environmental setups, and the problem of generalization is more apparent. That model is trained on our dataset which is automatically collected as detailed in subsection \ref{sec:model}, while the preceding two models are trained on data collected by a human driver. The ablated model trained without data augmentation performs worst compared to the other models which aligns with the conclusion in \cite{codevilla2018end} that careful data augmentation is crucial for generalization even within the training town. The last two models in the table ("Camera + LiDAR" and "Camera + LiDAR (Our Route Planner)") are also trained on our automatically-collected dataset.

\begin{table*}[!h] 
	\def\arraystretch{1.5}
	\centering
	\caption{The average number of kilometers traveled before an infraction}
	\label{table:collisions_in_different_tasks_results}
	\setlength{\tabcolsep}{0.4em} 
    \renewcommand{\arraystretch}{1.0}
	\begin{tabular}{|c|c||c|c|c|c|}
	\hline
	\multirow{3}{*}{Infractions}         	& \multirow{3}{*}{Model}	& \multicolumn{4}{c|}{Average kilometers traveled before an infraction}					\\ \cline{3-6}
											&							& \multicolumn{2}{c|}{Training town}	& \multicolumn{2}{c|}{New town} 						\\ \cline{3-6}
											&							& Training weather	& New weathers		& Training weather	& New weathers 	\\ \hline \hline
\multirow{4}{*}{\makecell{Collision\\to a\\Pedestrian}} & Camera only, \cite{dosovitskiy2017carla} pre-trained & 7.15 & \textbf{19.48} & 0.99 & 2.18 \\ 
& Camera only (trained on our data) & 6.08 & 5.32 & 1.49 & 3.14 \\ 
& Camera + LiDAR & 60.96 & 15.76 & 25.01 & 12.60 \\ 
& \textbf{Camera + LiDAR (Our Route Planner)} & \textbf{62.02} & 18.79 & \textbf{25.50} & \textbf{16.15} \\ \hline \hline
\multirow{4}{*}{\makecell{Collision\\to a\\Vehicle}} & Camera only, \cite{dosovitskiy2017carla} pre-trained & 1.35 & 0.89 & 0.18 & 0.17 \\ 
& Camera only (trained on our data) & \textbf{1.59} & \textbf{1.33} & 0.44 & 0.70 \\ 
& Camera + LiDAR & 1.13 & 1.09 & 0.51 & 0.84 \\ 
& \textbf{Camera + LiDAR (Our Route Planner)} & 1.36 & 1.20 & \textbf{1.83} & \textbf{2.02} \\ \hline \hline
\multirow{4}{*}{\makecell{Collision\\to a\\Static Object}} & Camera only, \cite{dosovitskiy2017carla} pre-trained & \textbf{5.50} & \textbf{5.56} & 0.29 & 0.85 \\ 
& Camera only (trained on our data) & 5.15 & 2.66 & 0.26 & 0.33 \\ 
& Camera + LiDAR & 3.05 & 3.15 & 0.54 & 0.90 \\ 
& \textbf{Camera + LiDAR (Our Route Planner)} & 3.94 & 3.70 & \textbf{1.03} & \textbf{1.24} \\ \hline \hline
\multirow{4}{*}{\makecell{Going\\Outside\\of Road}} & Camera only, \cite{dosovitskiy2017carla} pre-trained & 14.30 & 12.98 & 0.45 & 0.90 \\ 
& Camera only (trained on our data) & 11.15 & 7.97 & 0.59 & 0.79 \\ 
& Camera + LiDAR & 10.16 & 10.50 & 0.96 & 1.57 \\ 
& \textbf{Camera + LiDAR (Our Route Planner)} & \textbf{64.04} & \textbf{32.37} & \textbf{2.75} & \textbf{3.23} \\ \hline \hline
\multirow{4}{*}{\makecell{Invading the\\Opposite Lane}} & Camera only, \cite{dosovitskiy2017carla} pre-trained & 4.77 & 9.74 & 0.51 & 1.69 \\ 
& Camera only (trained on our data) & 13.38 & 15.95 & 0.77 & 0.90 \\ 
& Camera + LiDAR & 8.71 & 10.50 & 0.93 & 1.05 \\ 
& \textbf{Camera + LiDAR (Our Route Planner)} & \textbf{21.35} & \textbf{32.37} & \textbf{11.00} & \textbf{16.15} \\ \hline \hline
\multirow{1}{*}{Violating Traffic Light}
& Camera + LiDAR (Our Route Planner) & 57.43 & 53.05 & 32.88 & 28.92 \\ \hline \hline
	\end{tabular}
\end{table*}

The proposed model ("Camera + LiDAR" model) results demonstrate that it performs significantly better than the CIL model \cite{codevilla2018end} in every task and environmental setup combination, even while it is trained on driving data recorded automatically. Both models use the same global route planner algorithm in \cite{dosovitskiy2017carla} and \cite{codevilla2018end}. The learned driving policy consistency against varying weather conditions is improved by 3.91 times, as the average (per task and town) success rate difference due to changing from weathers seen during training to unseen weathers becomes 1.375\%, while it is 5.375\% in case of the "Camera only, \cite{dosovitskiy2017carla} results" model. On the hardest task of "Dynamic Navigation", the autonomous driving success rate is improved by 52\% when deployed in the new town and weathers that are unseen during training; the success rate improved from 42\% to 64\%. Succeeding in those "Dynamic Navigation" test scenarios requires responding to traffic lights. The model is able to recognize and respond to traffic lights as situations involving traffic lights are part of the training data. When our new global route planner is adopted instead, performance is further improved. The autonomous driving success rate is improved by 37\%; the success rate improved from 64\% to 88\%. Figure \ref{fig:route_planner} shows two example snapshots showing the global route planner during deployment. In the first example in figure \ref{fig:route_planner}(a), our planner returns a "follow lane" navigational command, while the planner adopted in \cite{codevilla2018end} returns a faulty "go left" command causing the vehicle to invade the opposite lane and crash with other vehicles.

\subsection{Infractions Analysis}
\label{subsec:infractions_results}

In table \ref{table:collisions_in_different_tasks_results}, we report the average number of kilometers traveled before an infraction for each model on the "Dynamic Navigation" task test scenarios, the higher the numbers the better. For the majority of infractions types, the proposed model performed better than the other models, especially in the new town unseen during training, which emphasizes the generalization improvement achieved by our model. The model is demonstrated to avoid both dynamic (pedestrian and other vehicles) and static objects. The improvement margin is further increased when our global route planner is adopted. Figure \ref{fig:route_planner}(a) shows an example justifying the significant improvement in the rates of invading the opposite lane and going outside the road when using the proposed global route planner. Models trained on data that include situations involving traffic lights are observed to achieve good results in responding to traffic lights. Overall, the infractions analysis is a strong indicator that further progress is still required to produce reliable and safe autonomous driving.

\subsection{Road Blockages Avoidance Benchmark}
\label{subsec:road_Blockages_avoidance_results}

\begin{figure}[h]
    \centering

    \begin{minipage}{1.0\linewidth}
        \centering
        \subfigure[]{\label{fig:ogm:scala}
        \fbox{
        \includegraphics[width=0.5\linewidth]{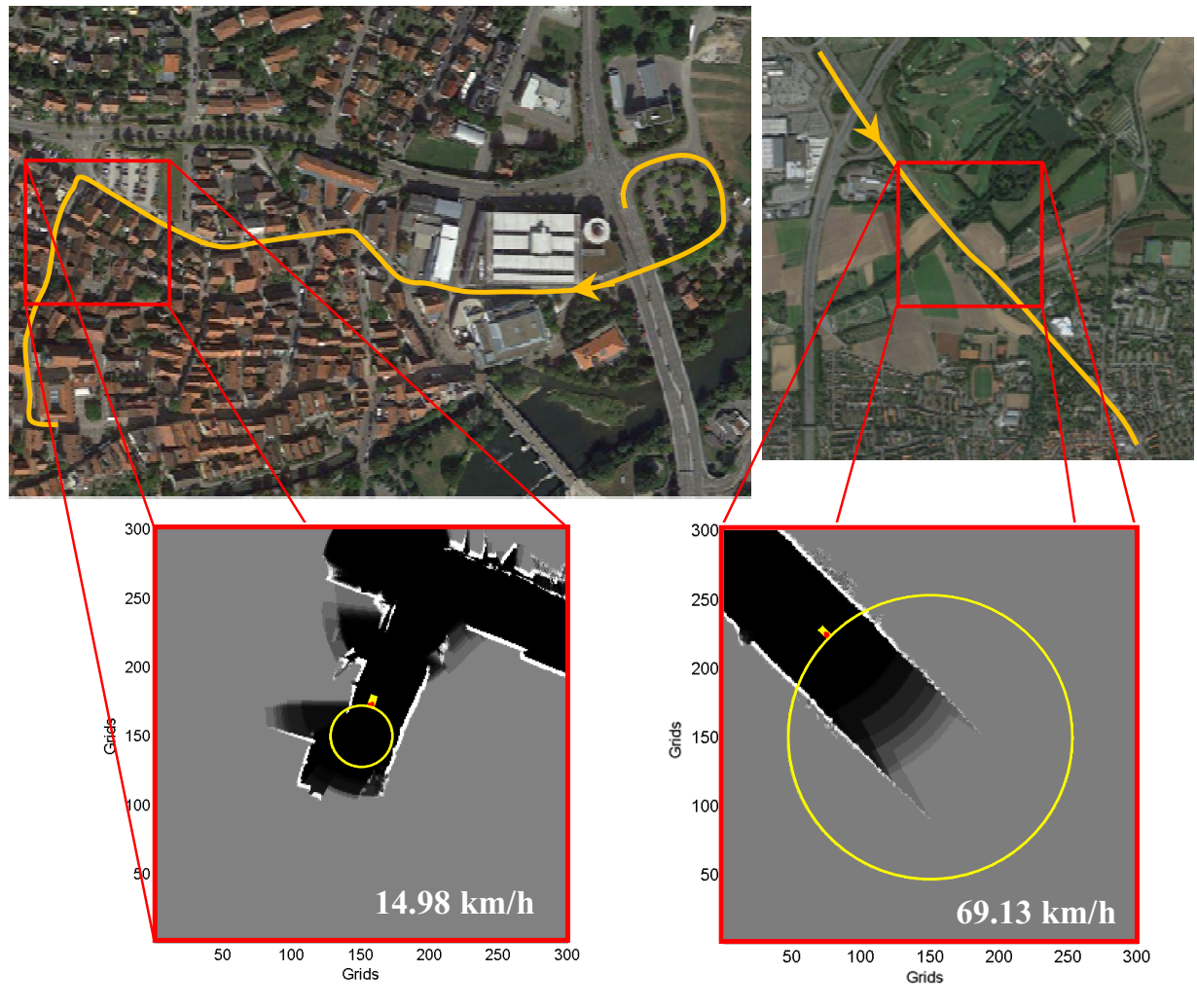}}
        }
    \end{minipage}
    
    \begin{minipage}{1.0\linewidth}
        \centering
        \subfigure[]{\label{fig:ogm:CARLA_samples}
        \fbox{
        \includegraphics[height=0.15\linewidth]{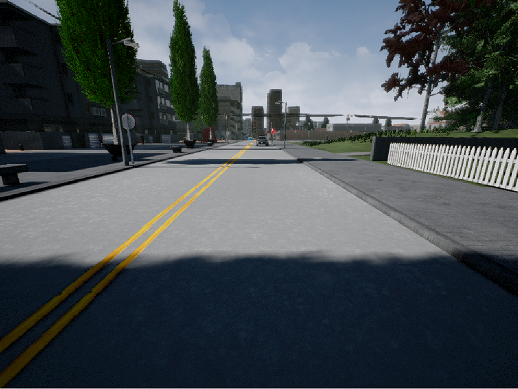}
        \includegraphics[height=0.2\linewidth]{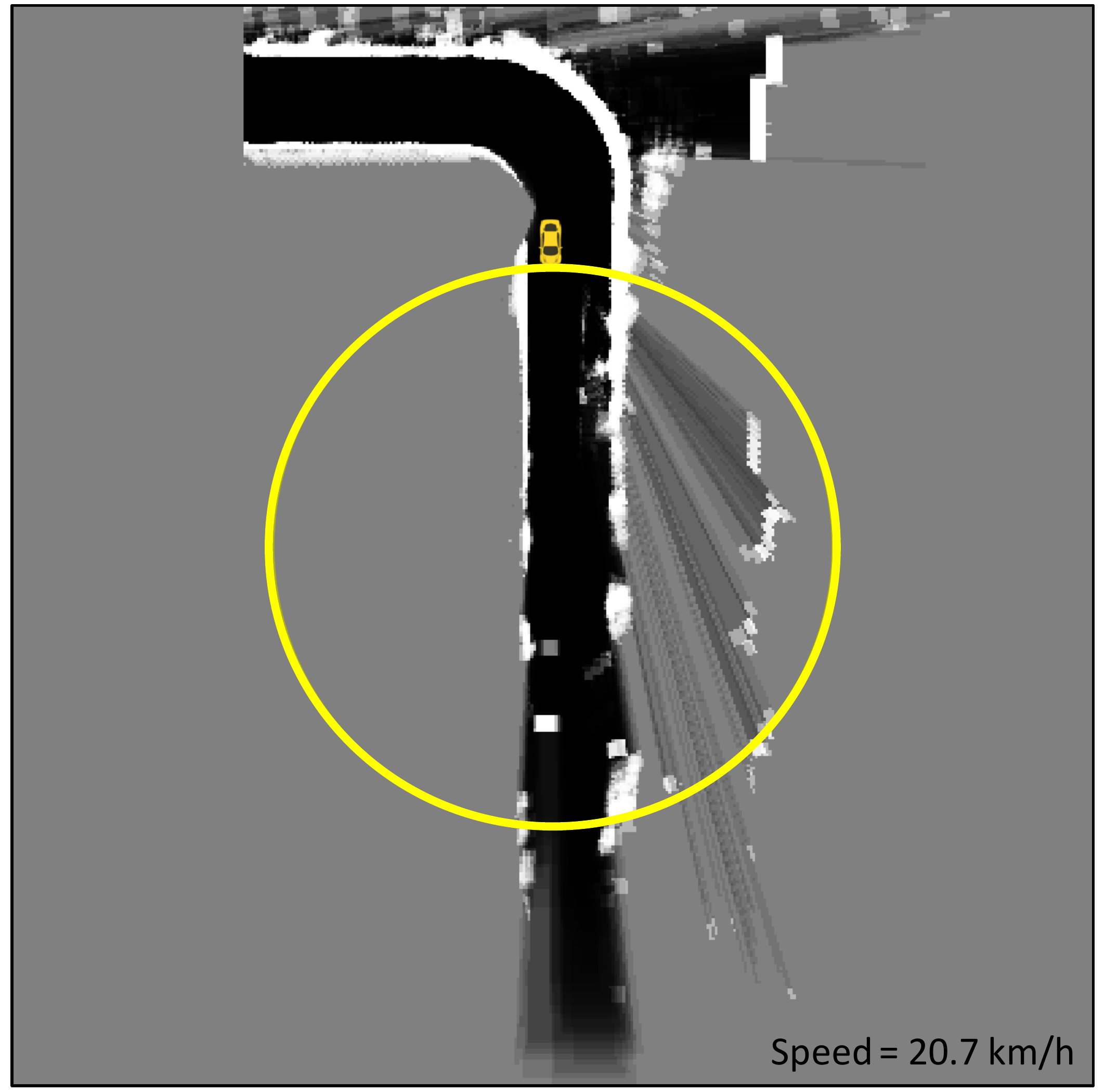}
        }

        \fbox{
        \includegraphics[height=0.15\linewidth]{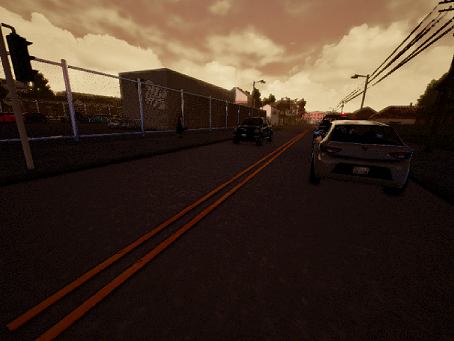}
        \includegraphics[height=0.2\linewidth]{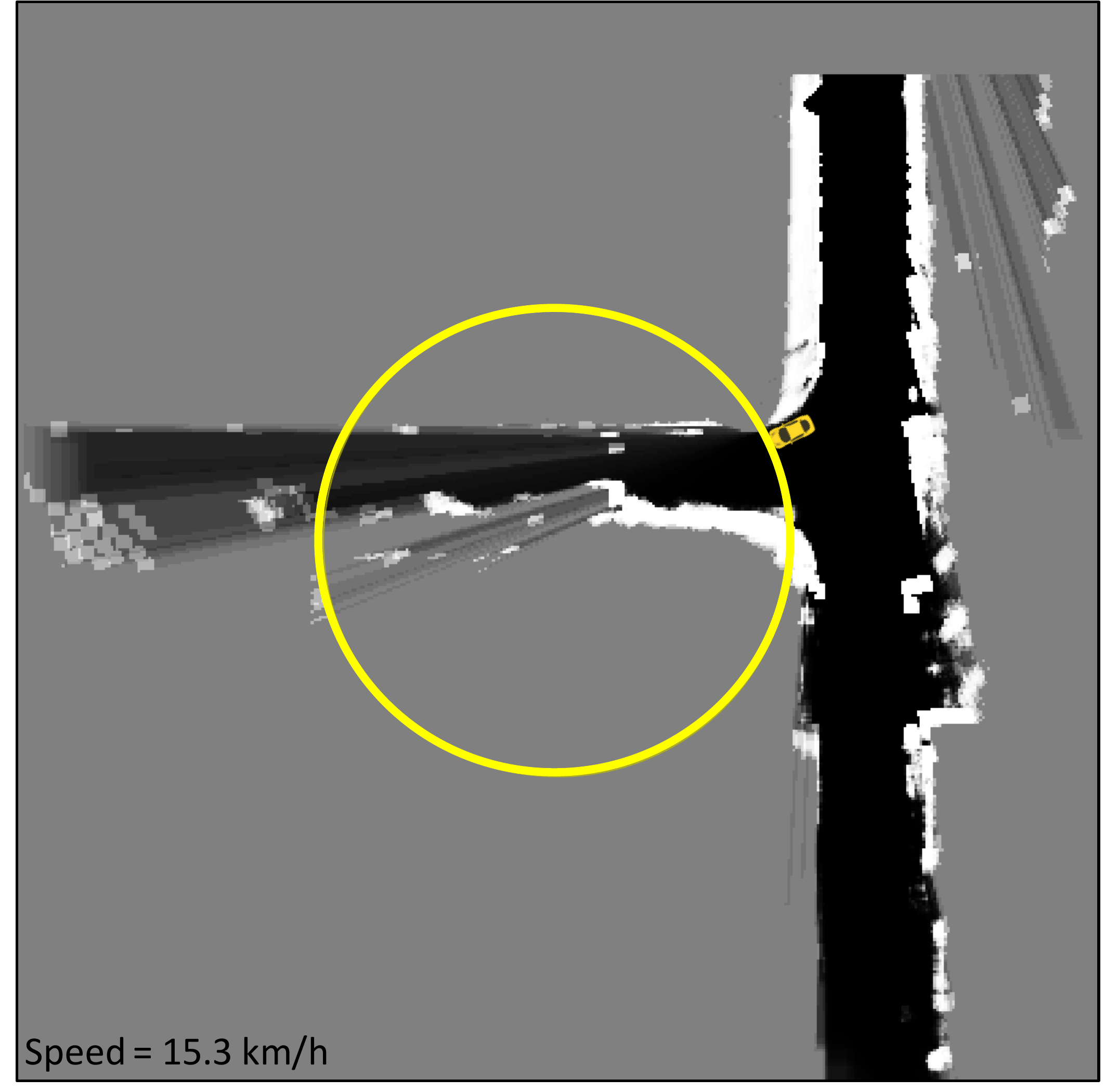}
        }}
    \end{minipage}

    
    \caption{Sample real-world and simulation results for our OGM method with the vehicle position circle shown in yellow. (a) Real-world results on urban and highway driving scenarios. (b) Results from CARLA simulator along with the corresponding RGB camera image.}
    \label{fig:ogm}
\end{figure}

Figure \ref{fig:ogm} shows sample real-world and simulation results for our proposed OGM method, with the vehicle position circle shown in yellow. For the real-world results, Valeo ScaLa first-generation LiDAR \cite{valeoscala} is used, which is the first laser scanner for automotive volume production. The figure shows two examples out of hours of testing. The laser scanner data for both of the two examples are recorded in Stuttgart, Germany. The first sample is for urban city driving and the second one is for high-way driving. Practically, we found that the \textit{get\_affected\_area} operation is around five times faster using the convex hull option on average. But the number of grid cells to update is around 2.37 times larger for such an option, which makes it overall around 1.5 times slower than the polygon option. The figure also shows two sample OGM results on CARLA simulator along with the corresponding RGB camera image. In Algorithm \ref{algo:ogm}, $log\_odd_{occ}$ and $log\_odd_{free}$ are tuned in the training town and set to 0.9 and 0.7 respectively. The resolution is set to 0.5 meters, $w$ is set to 1 meter, and $\alpha$ is set to $2\circ$. The samples shown in figure \ref{fig:ogm:CARLA_samples} are from the test town demonstrating generalization to new environments. The used 360-degree LiDAR has 32 layers, a vertical field of view from $-30\circ$ to $10\circ$, and a 150 meters range. CARLA LiDAR does not model beam echoes and their pulse width which are not required by our proposed PGV and OGM representations.

\begin{figure}[!t]
  \centering
  \includegraphics[width=9.0cm]{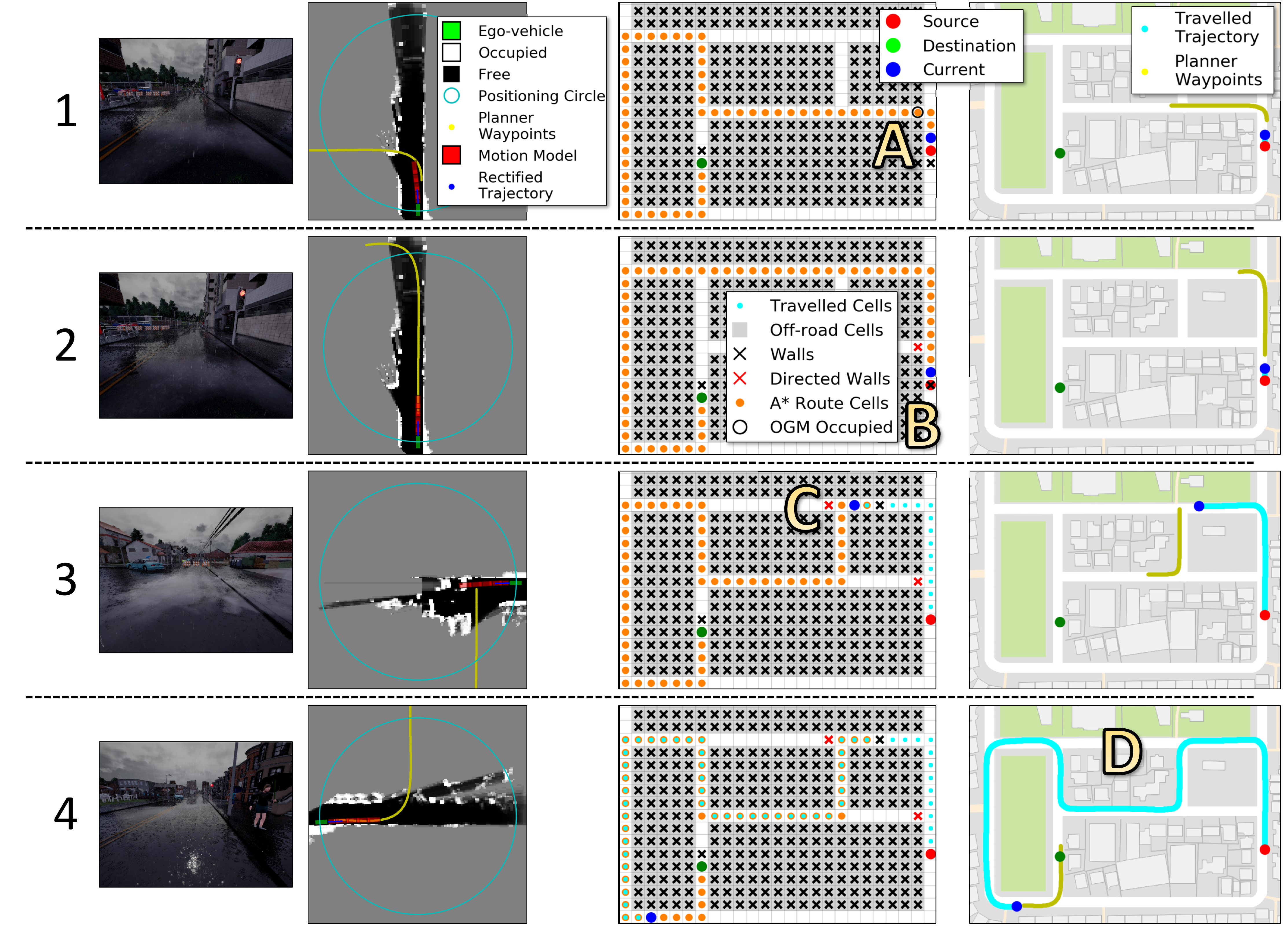}
  \caption{Snapshots from an example test scenario for the proposed system during deployment in a new town unseen during training. (A) shows the moment a road blockage is detected, then immediately after the vehicle is guided to follow another route to reach the destination as in (B). (C) shows another road blockage detected, and (D) shows the trajectory the vehicle followed by avoiding the two road blockages towards the destination to eventually arrive successfully.}
  \label{fig:road_blockages_avoidance_example_result_scenario}
\end{figure}

The CARLA urban driving benchmark \cite{dosovitskiy2017carla} hardest tasks of navigation and dynamic navigation are adapted to test road blockages avoidance. CARLA simulator is upgraded to support adding random road blockages by following two criteria. Firstly, each experiment in the benchmark has at least one road blockage and up to five along the shortest route towards the destination. The number of blockages is sampled from a uniform distribution. Secondly, in 50\% of the benchmark experiments, there is at least one blockage that mandates dynamically changing the route to reach the destination. In the other 50\% of the experiments, the road blockages along the route are partial and do not require rerouting, as an example for being in another lane. For the dynamic navigation task, the simulator autopilot logic is overridden to prevent other vehicles and pedestrians from colliding or being stuck in road blockages. The involved thresholds in the road blockages avoidance algorithm are tuned in the training town. In Algorithm \ref{algo:route_planner}, the parameters $res$, $far\_inters$, $inter\_exited$ are set empirically to 8.215 meters (as in \cite{codevilla2018end}), 4 cells, and 1 cell respectively. In Algorithm \ref{algo:road_blockages}, $route\_pts$ is set empirically to 5 cells given the used $res$ value in Algorithm \ref{algo:route_planner}, while the used OGM width and height are set to 80 meters and the resolution is set to 0.5 meters. The chosen values of those parameters are demonstrated to detect road blockages at around 60 meters ahead in most of the cases as in the examples in figures \ref{fig:navigation_loop} and \ref{fig:road_blockages_avoidance_example_result_scenario}. The OGM quality, which is directly affected by the weather conditions \cite{wallace2020full}, and occupancy confidence are the main factors determining the blockage detection range. In adverse weather conditions, the range is found reduced to around 30 meters.

Table \ref{table:road_blockage_benchmark_success_rates} reports the autonomous driving success rates with and without using the proposed road blockages avoidance method, and against the state-of-the-art CIL model \cite{codevilla2018end}. The road blockages avoidance algorithm improved the driving success rate by 27\% on average over all the benchmark tasks and conditions. Table \ref{table:road_blockage_benchmark_collisions_with_static_objects_results} reports the average number of kilometers traveled before a collision to a static object on the different condition for the hardest task of "Dynamic Navigation". The table confirms that the introduced road blockage algorithm significantly reduced infractions with static objects. The average kilometers traveled before a collision to a static object increased by more than 1.5 times.

\begin{table}[!h] 
	\centering
	\caption{Road blockages avoidance benchmark: driving success rate average percentage}
	\label{table:road_blockage_benchmark_success_rates}
	\setlength{\tabcolsep}{0.1em} 
    \renewcommand{\arraystretch}{1.2} 
	\begin{tabular}{|c|c||c|c|c|c|}
		\hline
		\multirow{3}{*}{Task}								& \multirow{3}{*}{Model}							            & \multicolumn{4}{c|}{\makecell{Percentages of average success rate}}	\\ \cline{3-6}
															&													            & \multicolumn{2}{c|}{Training town}	& \multicolumn{2}{c|}{New town} 					\\ \cline{3-6}
															&													            & \makecell{Training\\weathers}	    & \makecell{New\\weathers}	& \makecell{Training\\weathers}	& \makecell{New\\weathers} \\ \hline \hline

\multirow{5}{*}{Navigation} & Camera model in \cite{dosovitskiy2017carla} & 56 & 50 & 32 & 40 \\ \cline{2-6}
						    & Camera+LiDAR Model & 56 & 56 & 44 & 44 \\ \cline{2-6}
						    & \textbf{\makecell{Camera+LiDAR Model,\\with road blockages\\avoidance}} & \textbf{94} & \textbf{96} & \textbf{72} & \textbf{74} \\ \hline \hline

\multirow{5}{*}{\makecell{Dynamic\\Navigation}} & Camera model in \cite{dosovitskiy2017carla} & 55 & 48 & 29 & 32 \\ \cline{2-6}
						                        & Camera+LiDAR Model & 49 & 50 & 45 & 44 \\ \cline{2-6}
						                        & \textbf{\makecell{Camera+LiDAR Model,\\with road blockages\\avoidance}} & \textbf{79} & \textbf{78} & \textbf{57} & \textbf{56} \\ \hline \hline
	\end{tabular}
\end{table}

\begin{table}[!h] 
	\def\arraystretch{1.5}
	\centering
	\caption{Road blockages avoidance benchmark: the average number of kilometers traveled before a collision to a static object}
	\label{table:road_blockage_benchmark_collisions_with_static_objects_results}
	\setlength{\tabcolsep}{0.1em} 
    \renewcommand{\arraystretch}{1.2}
	\begin{tabular}{|c||c|c|c|c|}
	\hline
	\multirow{3}{*}{Model}	& \multicolumn{4}{c|}{\makecell{Average kilometers traveled\\before an infraction}}					\\ \cline{2-5}
							& \multicolumn{2}{c|}{Training town}	& \multicolumn{2}{c|}{New town} 						\\ \cline{2-5}
							& \makecell{Training\\weather}	& \makecell{New\\weathers}		& \makecell{Training\\weather}	& \makecell{New\\weathers} 	\\ \hline \hline
											
	Camera model in \cite{dosovitskiy2017carla} & 0.58 & 0.56 & 0.29 & 0.27 \\ \cline{1-5}
	Camera+LiDAR Model & 0.6 & 0.55 & 0.32 & 0.34 \\ \cline{1-5}
	\textbf{\makecell{Camera+LiDAR Model,\\with road blockages avoidance}} & \textbf{2.05} & \textbf{1.73} & \textbf{0.69} & \textbf{0.52} \\ \hline \hline
	\end{tabular}
\end{table} 

Figure \ref{fig:road_blockages_avoidance_example_result_scenario} shows an example scenario for our system during deployment in the test town. The ego-vehicle detects two road blockages and dynamically estimates and follows new routes to eventually reach the designation successfully. On average over all the benchmark test scenarios, the overhead of adding the LiDAR to the system, including the PGV and the OGM processing and increasing the model size due to adding the LiDAR input modality, increases the overall system runtime by 68.83\%.

\section{Conclusion}
\label{sec:conclusion} 

We proposed a model that extends the state-of-the-art conditional imitation learning method by fusing a LiDAR sensor input with the camera aiming to tackle the challenges of lack of generalization and inconsistency against varying weather conditions. Additionally, we introduced a new efficient Occupancy Grid Mapping method that improves runtime performance, memory utilization, and map accuracy. The OGM is used to upgrade the conditional imitation learning method to dynamically detect partial and full road blockages and guides the controlled vehicle to another route to reach the destination. On CARLA simulator urban driving benchmark, camera and LiDAR fusion is demonstrated to improve weather consistency by around four times. The model has shown to significantly improve the autonomous driving success rate and average distance traveled towards the destination on all the driving tasks and environments combinations while being trained on automatically recorded data. The generalization to new environments in terms of driving success rate has improved by 52\%. The infractions analysis showed improvement as well but overall indicates that further progress is still required to produce reliable and safe autonomous driving.

The global route planner provided more accurate navigational commands and improved the driving success rate further by 37\%. CARLA benchmark is upgraded to allow test navigation while having unexpected temporary stationary road blockages. Our road blockages avoidance algorithm improved the driving success rate by 27\% and the average kilometers traveled before a collision to a static object increased by more than 1.5 times.

In future work, we need to investigate the proposed model's capability to drive on multi-lane roads and through roundabouts. In addition, knowing whether the generalization issue is caused more by the different road layout or the different environment domains and textures can guide to further improve the model generalization. Moreover, the PGV allows standard CNN to achieve real-time performance; however, the discretization errors can be mitigated by adopting sparse convolution which requires studying the speed-accuracy trade-off.







\bibliographystyle{IEEEtranS}
\bibliography{references}

\end{document}